\documentclass[11pt]{article}
	
	\newcommand{\blind}{0}
	
    \makeatletter
    \renewcommand\section{\@startsection {section}{1}{\z@}%
                                       {-3.5ex \@plus -1ex \@minus -.2ex}%
                                       {2.3ex \@plus.2ex}%
                                       {\normalfont\fontfamily{phv}\fontsize{16}{19}\bfseries}}
    \renewcommand\subsection{\@startsection{subsection}{2}{\z@}%
                                         {-3.25ex\@plus -1ex \@minus -.2ex}%
                                         {1.5ex \@plus .2ex}%
                                         {\normalfont\fontfamily{phv}\fontsize{14}{17}\bfseries}}
    \renewcommand\subsubsection{\@startsection{subsubsection}{3}{\z@}%
                                        {-3.25ex\@plus -1ex \@minus -.2ex}%
                                         {1.5ex \@plus .2ex}%
                                         {\normalfont\normalsize\fontfamily{phv}\fontsize{14}{17}\selectfont}}
    \makeatother
	
	\usepackage{amsmath}
	\usepackage{graphicx}
	\usepackage{enumerate}
	\usepackage{xcolor}
	\usepackage{natbib} 
	\usepackage{url} 
    \usepackage{hyperref}
	
    \usepackage{xcolor} 
    \usepackage{comment}
    \usepackage{subcaption}
    \newtheorem{definition}{Definition}
    
    \newtheorem{remark}{Remark}
    \newtheorem{proposition}{Proposition}
    \usepackage{algpseudocode}
    \usepackage{algorithm}
    \usepackage{enumitem}
    \usepackage{multirow}
    \newcommand{\ignore}[1]{}
    \usepackage{makecell}
    \usepackage{longtable}
    \newcolumntype{P}[1]{>{\centering\arraybackslash}p{#1}}

\begin{document}
		
		\def\spacingset#1{\renewcommand{\baselinestretch}%
			{#1}\small\normalsize} \spacingset{1}
		
		\if0\blind
		{
			\title{\bf \emph{A Consistency-Centric Approach to Set-Based Optimization with Multiple Models of Unranked Fidelity}}
			\author{ Danielle F. Morey $^a$, Giulia Pedrielli $^b$, Cherry Y. Wakayama $^c$, and Zelda B. Zabinsky $^a$ \\\\
			$^a$ Industrial and Systems Engineering, University of Washington, Seattle WA, USA \\
             $^b$ School of Computing and Augmented Intelligence, \\Arizona State University, Tempe AZ, USA \\
             $^c$ Intelligence, Surveillance, and Reconnaissance Department, \\Naval Information Warfare Center Pacific, San Diego CA, USA}
			\date{}
			\maketitle
		} \fi
		
		\if1\blind
		{

            \title{\bf \emph{IISE Transactions} \LaTeX \ Template}
			\author{Author information is purposely removed for double-blind review}
			
\bigskip
			\bigskip
			\bigskip
			\begin{center}
				{\LARGE\bf \emph{IISE Transactions} \LaTeX \ Template}
			\end{center}
			\medskip
		} \fi
		\bigskip

    \begin{abstract}

    In complex real-world settings, optimization is challenged by the presence of diverse models of differing fidelity. 
    In many optimization problems, a single model is treated as the most accurate representation of the underlying system, while other models are evaluated primarily by their agreement with this presumed most accurate model. Yet in real-world applications, model accuracy is rarely known a priori and assuming a single most accurate model can be misleading. This paper addresses this gap by proposing a flexible set-based optimization methodology called Set-Based Optimization with Multiple Models (S-BOMM) that works with multiple models without the assumption of a most accurate high-fidelity model. Unlike traditional optimization approaches that focus on finding an optimal solution according to the high-fidelity model, our methodology utilizes consistency between models to identify good solutions across multiple models. 
    A probabilistic analysis of the consistency method is provided that bounds the likelihood of the methodology producing correct or incorrect results. Empirical results demonstrate the effectiveness of S-BOMM on test problems. 
    By focusing on the consistency across models rather than relying on a single best solution, this set-based approach offers a practical alternative to optimization problems where multiple models must be considered without assuming a single most accurate high-fidelity model.

	\end{abstract}
			
	\noindent%

	\spacingset{1.5} 

\section{Introduction}

In complex optimization problems where multiple models represent different perspectives and/or fidelities of a system, 
a structured approach is needed to leverage model consistency for decision-making.
Typically, multi-fidelity optimization methods assume a single model is most accurate (e.g., a detailed simulation) and considered high-fidelity, while other models (e.g., queueing network models) are assumed to be less accurate and considered low-fidelity~\citep{LiLi2024,Peherstorfer2018}. 
Ensembling methods, such as random forests, are commonly used in machine learning and other data-centric techniques to combine multiple models~\citep{Opitz1999_ensemble,Parmar2018_ensemble}
In reality, it is often true that no single model provides an accurate and complete picture of the complex system it is designed to emulate. This is especially prevalent in engineering design problems that lack an existing physical system~\citep{Bramerdorfer2018,Zabinsky_2006_composite}.
Without a known most accurate model, no single solution can be deemed optimal, as competing models may yield conflicting optima with no definitive basis for deciding among them.
Rather, it is more useful to identify a set of solutions that are consistently evaluated as high quality across multiple models. 

To effectively apply 
optimization in the absence of a ranked hierarchy of fidelity, it is crucial to compare solutions across multiple models. 
We present an approach that operates on subregions of the solution space because consistency (agreement across models) emerges more reliably at the aggregate level than at individual point solutions.
By classifying subregions of solutions based on observed function values, a set of solutions can be identified that consistently classifies in a desirable manner across multiple models.  
The decision-maker may have a goal of identifying solutions within the best 10\% of solutions. In this way, the target region would the 10th percentile of solutions. An example of three classes that support this goal are: a subregion resides entirely within the target region, resides entirely outside the target region, or overlaps the target region.
This classification can then be compared across multiple models to determine which subregions consistently classify in this way.  Consistency of classification across multiple models thus provides a practical basis for identifying promising subregions without relying on a single model as most accurate.
If most or all models agree that a subregion of the decision space is promising, we can confidently mark that subregion as promising. Likewise, if most or all models agree that a subregions is not promising, we can stop spending (or reduce the amount of) computational effort in that region. In regions where there is little to no consistency between models, it is advisable to reevaluate each model to determine if there are opportunities for improving accuracy and consistency. 

In this paper, we introduce Set-Based Optimization with Multiple Models (S-BOMM), a methodology designed for 
multiple model optimization where model accuracy is unknown.
Additionally, S-BOMM supports set-based optimization, making it well suited for problems where system variability and implementation constraints favor finding a set of ``good" solutions rather than a single optimum.

\subsection{\emph{Related Work}}\label{s:background}
In optimization of complex systems, available models often span differing fidelities and computational effort, from analytical models that make simplifying assumptions to computationally intensive detailed simulations.

Every model comes with its own set of assumptions and limitations that must be considered, as well as the computational expense of that model. Models with minimal assumptions often come at high computational expense. These are often referred to as high-fidelity models and treated as accurate representations of the real system. Examples include digital twins~\citep{Alam2017,Sel2025,Tan02062024,Zhang2022}, agent-based modeling~\citep{Macal_AgentSim,Sanchez2002,Yin2024}, and discrete event simulations~\citep{Banks1986_DES,Choi2017}. 
To alleviate the computational expense of these models, decision makers often utilize computationally cheaper models.
Computationally cheaper models can be built using knowledge of the system and making simplifying assumptions, such as queueing models~\citep{filipowicz2008queueing,Liu2021,TOMO_WSC2021,TOMO_Frontiers} and  compartmental models~\citep{Lee2021,Tolles2020,walter1999compartmental}. Surrogate models, or metamodels, can be developed from the computationally expensive model~\citep{Barton1994,Barton2006,Conn2009,Simpson2001}, including Gaussian processes~\citep{Costabal2019,Gnanasambandam2025,MacKay1998_GP}, multi-task Gaussian processes~\citep{Li2018,Zhang2010}, and polynomial regression~\citep{Cao2025_RPR,Jaenisch2011_RPR,Wan_2019_RPR}.

Intuitively, leveraging multiple models of a system allows one to capitalize on the strengths of each model while mitigating their individual limitations. The challenge then becomes determining how to effectively 
utilize multiple models
in a way that allows for analysis and optimization. 
One approach to utilizing low- and high-fidelity models is to rank order solutions in the decision space based on low-fidelity function evaluations and then evaluate the highest ranking solutions with the high-fidelity model~\citep{Xu2015,Xu2016}.
\cite{Zabinsky2019} expanded upon multi-fidelity literature by determining in what regions a low-fidelity model was accurate enough to use in place of the high-fidelity model. These methods, however,  are still based on how well the low-fidelity model outputs matched with those of the high-fidelity model.

For some complex systems, however, it may be inadequate to assume one model has higher accuracy than other models. \cite{TOMO_Frontiers} created a high-fidelity simulation for the optimal design of a data collection scheme using unmanned underwater vehicles (UUVs) in an underwater sensor network, as well as a low-fidelity analytic queueing model. During the analysis of these two models, it was found that the low- and high-fidelity models were not agreeing on system performance. Upon closer inspection, one key assumption in the low-fidelity model, namely the assumption of load balancing, was found to be the main contributor to this inconsistency. 
The high-fidelity simulation did not rely on the assumption of load balancing, instead it used shortest processing time for scheduling.
Rather than simply treating the high-fidelity simulation model as more accurately representing the system, a subject matter expert was consulted to determine the behavior of the system being modeled. 
In doing so, it was determined that a real-world scheduling scheme would likely implement a protocol to address load balancing. In this instance, the inconsistency between the low- and high-fidelity models led to model improvements on the high-fidelity model based on the insights gained from the low-fidelity model.
This is just one of many possible ways in which a model considered high-fidelity may not, in fact, be more accurate than other models. 
In this example, reconciling the differences between the simulation and the queueing model led to  model improvement and insights into desirable system performance.

Complex models may also suffer from model heterogeneity.  
Model heterogeneity refers to the variation in model accuracy across different subregions of the decision space. 
For example, models of  complex physical systems are often simplified by grid generation, where the model tends to perform well near the center of the grid but may become inaccurate near the border~\citep{aarnes2007_CoarseGrid}.

Another significant yet frequently overlooked challenge is the implementation of an optimal solution. An optimal solution may be difficult or even impractical to implement~\citep{Bramerdorfer2018,Xia2010,Zabinsky_2006_composite}. 
In such systems, finding a set of ``good" solutions is often more valuable than 
spending the additional time and computational effort required to identify a single optimum, especially when expert judgment significantly influences decision-making or when implementation may not be realistically attainable in practice.
Set-based optimization seeks to identify a set of optimal or near-optimal solutions~\citep{Zyl2023, PBnB_Chapter}. 
Partitioning methods, such as Probabilistic Branch and Bound~(PBnB)~\citep{Linz2017,PBnB_Chapter} and Part-X~\citep{PartX}, iteratively branch the decision space into subregions in order to identify the region or regions that provide good performance.
Set-based methods are especially advantageous in contexts where near-optimal solutions offer practical benefits not available with a single optimum.

\subsection{\emph{Contribution}}

In order to handle optimization problems with multiple models that lack an established hierarchy of accuracy, we introduce a model consistency algorithm that uses information from multiple models to gain insights about 
regions of good performance.
The Set-Based Optimization with Multiple Models (S-BOMM) methodology is designed to identify consistency across multiple models, providing a structured way to quantify agreement among models to guide the search for good solutions in complex systems.

This article makes four main contributions:
\begin{enumerate}
    \item Defines a new probabilistic-based consistency score to determine regions of consistent classification among multiple models.
    \item Establishes a methodology that embeds the consistency score within set-based optimization to aid decision making with multiple models.
    \item Provides an analysis that establishes the probability of correct and incorrect outcomes 
    and a discussion of the impact of parameter values.
    \item Provides two numerical examples to demonstrate that S-BOMM identifies a consistent solution set across multiple models.
\end{enumerate}

\subsection{\emph{Organization}}
The remainder of the paper is organized as follows. 
Section~\ref{s:SBOMM_overview} introduces the new consistency score and proposed S-BOMM methodology. 
Section~\ref{s:analysis} provides an analysis
that establishes
the probability of correct or incorrect consistent classification with an example illustrating the impact of parameter values.
Section~\ref{s:Empirical_Results} includes empirical results. 
Finally, Section~\ref{s:conclusion} provides concluding remarks, including implementation recommendations and future research directions. 

\section{Set-Based Optimization with Multiple Models (S-BOMM)}\label{s:SBOMM_overview}
The proposed methodology, S-BOMM, is designed to integrate multiple models  for decision making when a single, most accurate model is not available.
In the context of S-BOMM, a ``model" is 
an objective that can be evaluated on the bounded decision space.
Models may include, but are not limited to, a simulation, neural network,  and analytical model.
The proposed methodology is not restricted to any fixed model type and as such 
treats models as black boxes returning a function evaluation when executed at a specific location in the decision space.
S-BOMM is a partition-based methodology that relies on comparison of subregion classifications across different models to identify the most promising regions. 
While we make no assumption on the type of classifier, the classifier used must be the same across the available models.

\begin{figure}
    \centering
    \includegraphics[width=.75\linewidth]{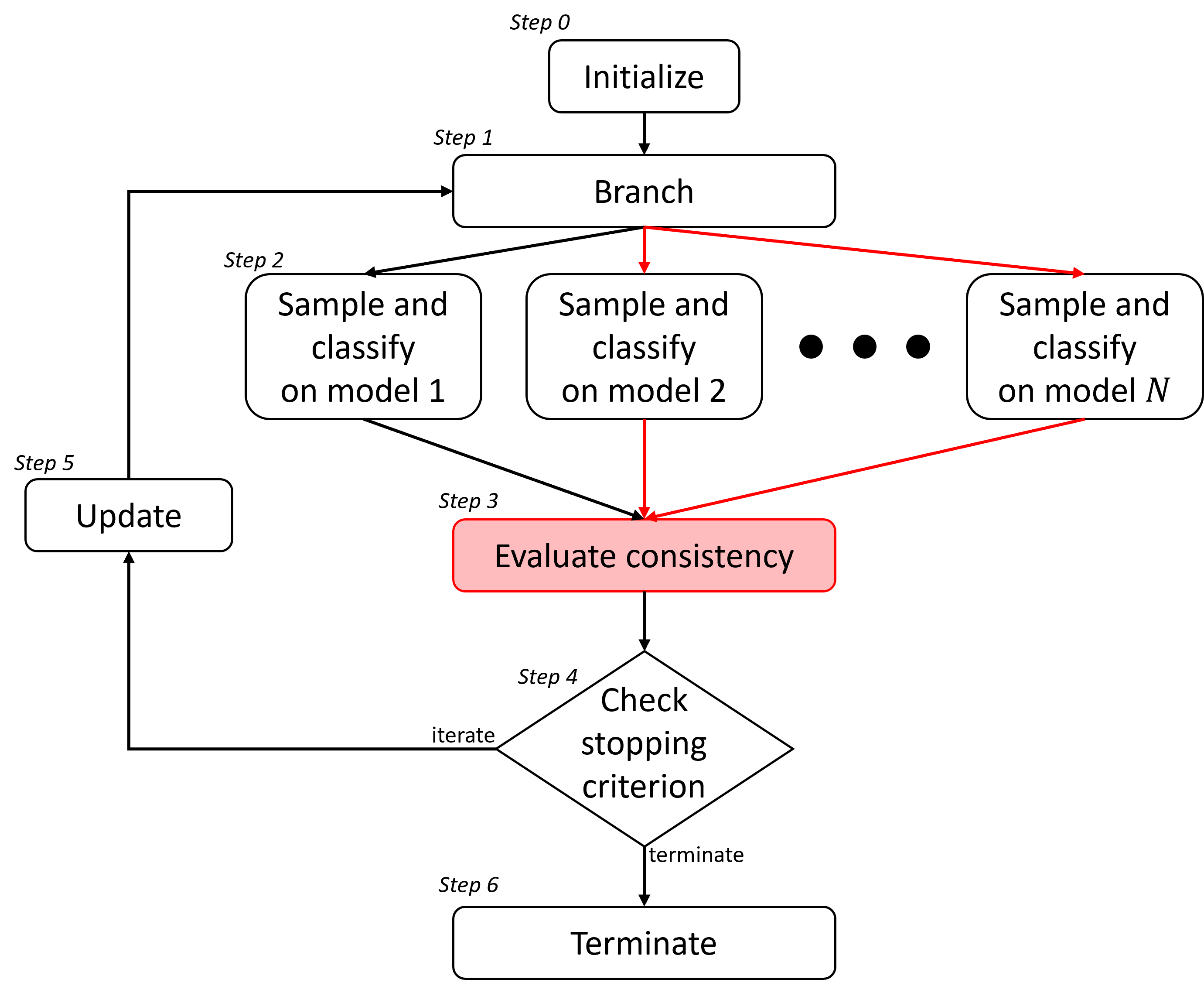}
    \caption{S-BOMM methodology.}
    \label{fig:S-BOMM_flowchart}
\end{figure}

An overview of the methodology is presented in Figure~\ref{fig:S-BOMM_flowchart} with the main contributions of this work highlighted in red. Namely, S-BOMM contributes to the literature by utilizing multiple models (indicated by the red arrows in Figure~\ref{fig:S-BOMM_flowchart}) and employing a consistency score (indicated by the red box in Figure~\ref{fig:S-BOMM_flowchart}) to integrate the classification of subregions across different models. 
S-BOMM can accommodate any user-selected branching scheme (Step 1) and user-selected sampling and classification  method (Step 2). 
The consistency score, presented later in this section, 
is used to determine which regions are consistently classified across models (Step~3).
Step 4 checks the user-defined stopping criterion. 
For the implementation in Section~\ref{s:Empirical_Results}, S-BOMM stops when $10\%$ of the domain space has been consistently classified as residing within the target region.

Before further detailing the methodology, we present the relevant notation.
Let $X$ be the bounded decision space. Let there be $N$ different models, $f_n(x)$ with $x\in X$ and $n=1,\ldots,N$.
When branching, let $\sigma_i$ be the $i$-th subregion in $X$, where the subregions have no intersection and cover the entirety of the decision space, that is, $\sigma_i\cap\sigma_j=\emptyset$ for all $i\neq j$ and $\bigcup_i\sigma_i=X$. 

Given~$K$ classes, 
a user-selected classifier assigns one of the $K$ possible classes to subregion $\sigma_i$ according to model~$n$.

\begin{definition}\label{def:Sn}
    Let $Y_n(\sigma_i)$ be the class $k\in\{1,\ldots,K\}$ assigned to subregion $\sigma_i$ when using the output of model $n$, $f_n(x)$, to evaluate $x\in\sigma_i$. 
\end{definition}
Notice that there are $K^N$ possible combinations of each of the $N$ models classifying as one of the $K$ classes. 
We refer to an element of the $K^N$ possible combination as a \textit{case} with $\mathcal{N}^t_{k}(\sigma_i)$ representing the set of models $n\in\{1,\ldots,N\}$ within case $t$ that result in the classification $Y_n(\sigma_i)=k$.
Case $t$ is the collection $\{\mathcal{N}_1^t(\sigma_i),\ldots,\mathcal{N}_K^t(\sigma_i)\}$ with $t\in\{1,\ldots, K^N\}$.

Let $p_n^k(\sigma_i)=P(Y_n(\sigma_i)=k)$ 
be the probability that subregion $\sigma_i$ is assigned to class~$k$ 
under model $n$.
It should be noted that $p_n^k(\sigma_i)$ may be impossible or impractical to evaluate directly and may require estimation.
For the implementation in Section~\ref{s:Empirical_Results}, $p_n^k(\sigma_i)$ is estimated using the confidence intervals described in Section~\ref{s:Empirical_Results}.

For each subregion, $\sigma_i$,k we define the consistency score across models given a classifier.
\begin{definition}\label{def:wc}
Given case $\{\mathcal{N}_1^t(\sigma_i),\ldots,\mathcal{N}_K^t(\sigma_i)\}$ with $t\in\{1,\ldots,K^N\}$, the consistency score 
for subregion $\sigma_i$ with respect to class $k$ is 
\begin{equation}\label{eq:weight}
    C_k^t(\sigma_i) = \sum_{n\in \mathcal{N}^t_{k}(\sigma_i)}p_n^k(\sigma_i)
\end{equation}
where by definition $\mathcal{N}^t_{k}(\sigma_i)$ is the set of models $n\in\{1,\ldots,N\}$ with $Y_n(\sigma_i)=k$ for $k\in\{1,\ldots,K\}$. 
\end{definition}

Given the consistency score, we can now define consistent classification for each subregion $\sigma_i$.

\begin{definition}\label{def:cc}
    Given case $\{\mathcal{N}_1^t(\sigma_i),\ldots,\mathcal{N}_K^t(\sigma_i)\}$  with $t\in\{1,\ldots,K^N\}$ and parameters $v$ and $r$ where $0< v\le N$ and $0< r\le N$,    subregion $\sigma_i$ is consistently classified as class 
    $k$ for $k\in\{1,\ldots,K\}$ iff 
\begin{equation}\label{eq:cc_v}
    C^t_k(\sigma_i)\ge v
\end{equation}
    and
\begin{equation}\label{eq:cc_r}
    C^t_k(\sigma_i) - C^t_j(\sigma_i) \ge r~\text{for each } j\in\{1,\ldots,K\},~j\neq k.
\end{equation}
    Subregion $\sigma_i$ is inconsistently classified iff (\ref{eq:cc_v}) and (\ref{eq:cc_r}) do not hold for any $k$
\end{definition}

Given case $\{\mathcal{N}^t_1(\sigma_i),\ldots,\mathcal{N}^t_K(\sigma_i)\}$, evaluate~(\ref{eq:weight}) to obtain consistency scores $C_1^t(\sigma_i)$ for $k=\{1,\ldots,K\}$. Then for each $k=\{1,\ldots,K\}$, evaluate (\ref{eq:cc_v}) and (\ref{eq:cc_r}) to determine if subregion $\sigma_i$ is consistently classified as class $k$.
A property of the definition of consistent classification is that, 
if a subregion is consistently classified as $k$, then this classification is unique. 
This is because, even if there are multiple $k$ that satisfy (\ref{eq:cc_v}), it is impossible for multiple $k$ to satisfy (\ref{eq:cc_r}) since $r>0$.
If there is no $k\in\{1,\ldots,K\}$ for which $\sigma_i$ is consistently classified, then $\sigma_i$ is inconsistently classified.

The parameters $v$ and $r$ serve as adjustable thresholds that determine when a region is considered consistently classified. 
Low values of both parameters correspond to a ``lenient" scheme, requiring only minimal agreement between models to consistently classify subregions. Increasing $v$ imposes a stricter requirement on the number of models that must agree with sufficiently high consistency score. Increasing $r$ imposes a stricter requirement on the difference between the consistency score of the consistent class and the consistency scores of the other classes.
Increasing $v$ and $r$ may lead to the inability to consistently classify subregions.

The output of S-BOMM is a solution set of subregions and how they are consistently classified. 
\begin{definition}
    A solution at termination of S-BOMM provides for each subregion $\sigma_i\in X$, whether it is consistently classified as class $k$ for any $k\in\{1,\ldots,K\}$ or is inconsistently classified.
\end{definition}

The S-BOMM methodology, as depicted in Figure~\ref{fig:S-BOMM_flowchart}, follows. 

\begin{enumerate}[start=0,label=\textbf{Step \arabic*}:, leftmargin=*, align=left]
    \item \textbf{Initialize.} Initialize each model, $n\in\{1,\ldots,N\}$ over the decision space, $X$. Define classes $\{1,\ldots,K\}$ and set parameter values $v$ and $r$. Initialize all parameters necessary for the user-selected branching scheme and user-selected sampling and classification method.
    \item \textbf{Branch.} Branch the decision space, $X$, according to the user-selected branching scheme, resulting in a set of subregions, $\{\sigma_i\}\subseteq X$. 
    \item \textbf{Sample and Classify.} Utilize the user-selected sampling method to select sample points on the decision space and evaluate each sampled point according to model $n$, $f_n(x)$ for $x\in X$ and $n=\{1,\ldots,N\}$. Utilize user-selected classifier to assign a class to each subregion, $Y_n(\sigma_i)$ for each $\sigma_i\subseteq X$, according to model $n$. The classification of each subregion, $\sigma_i$, determines the case, $\{\mathcal{N}_1^t(\sigma_i),\ldots,\mathcal{N}_K^t(\sigma_i)\}$, for this iteration.
    \item \textbf{Evaluate Consistency.} For each subregion $\sigma_i$ and each class $k\in\{1,\ldots,K\}$, evaluate the consistency score, $C^t_k(\sigma_i)$, using to (\ref{eq:weight}). This may require estimating $p_n^k(\sigma_i)$. Then, for each $k$,  evaluate~(\ref{eq:cc_v})~and~(\ref{eq:cc_r}) and determine if subregion $\sigma_i$ is consistently classified as $k$ or is inconsistently classified.~\label{step:consistency} 
    \item  \textbf{Check Stopping Criterion.} If user-defined stopping criterion has been met, go to Step~6.  
    Otherwise, continue to Step~5.   
    \item \textbf{Update.} Use observations to update parameters needed for branching, sampling, and classification for all models, and  go to Step~1.  
    \item \textbf{Terminate.} Terminate and return the solution set of consistently classified subregions and inconsistently classified subregions.\label{step:terminate}
\end{enumerate}

An implementation of S-BOMM is described in Section~\ref{s:Empirical_Results}. In the next section, we explore the performance of the S-BOMM algorithm analytically.

\section{Analysis}\label{s:analysis}
We aim to determine the probability of consistently classifying a subregion correctly, consistently classifying a subregion incorrectly, or inconsistently classifying. To do this, we must first establish what it means to be ``correct", in the absence of a known hierarchy of accuracy among models.

Let $k_n^{\text{T}}(\sigma_i)$ be the true class of subregion $\sigma_i$ under model $n$, where $k_n^{\text{T}}(\sigma_i)$ may differ for each model $n$. An ideal classifier will satisfy $Y_n(\sigma_i)=k_n^{\text{T}}(\sigma_i)$ and $p_n^{k_n^{\text{T}}(\sigma_i)}=1$.  
A consistent classification is considered ``correct" when (\ref{eq:cc_v}) and (\ref{eq:cc_r}) are satisfied under an ideal classifier.
Let $k^*(\sigma_i)$ be the correct consistent classification.

To continue developing the probability of consistently classifying a subregion correctly, we express the probability of each of the $K^N$ possible cases occurring. Each element of a case, $\mathcal{N}_k^t(\sigma_i)$, is determined by the realization of $Y_n(\sigma_i)$ for all $n$, $n\in\{1,
\ldots,N\}$.

Assuming independence between models, the probability of case $t$ occurring is 
\begin{equation}\label{eq:prob_case}
    P(\{\mathcal{N}^t_1(\sigma_i),\ldots,\mathcal{N}^t_K(\sigma_i)\})=\prod_{k=1}^K \prod_{n\in\mathcal{N}^t_k(\sigma_i)}P(Y_n(\sigma_i)=k)=\prod_{k=1}^K \prod_{n\in\mathcal{N}^t_k(\sigma_i)}p_n^k(\sigma_i).
\end{equation}

In order for a case, $\{\mathcal{N}^t_1(\sigma_i),\ldots,\mathcal{N}^t_K(\sigma_i)\}$, to result in subregion $\sigma_i$ being consistently classified correctly, the associated consistency scores, $C_k^t(\sigma_i)$ for $k\in\{1,\ldots,K\}$, must satisfy (\ref{eq:cc_v}) and (\ref{eq:cc_r}) for $k^*(\sigma_i)$. 

\begin{remark}
    The probability of of consistently classifying subregion $\sigma_i$ correctly is
    \begin{equation}\label{eq:prob_corr}
    \begin{split}
        P(\sigma_i\text{ is consistently}&\text{ classified correctly as $k^*(\sigma_i)$}) \\
        &= \sum_{t:\{\mathcal{N}^t_1(\sigma_i),\ldots,\mathcal{N}^t_K(\sigma_i)\}\in\mathcal{A}(k^*(\sigma_i),\sigma_i)}
        ~~\prod_{k=1}^K ~~\prod_{n\in\mathcal{N}^t_k(\sigma_i)}p_n^k(\sigma_i)
    \end{split}
    \end{equation}
    where $\mathcal{A}(k^*(\sigma_i),\sigma_i)$ is the set of cases, $\{\mathcal{N}^t_1(\sigma_i),\ldots,\mathcal{N}^t_K(\sigma_i)\}$, $t\in\{1,\ldots,K^N\}$, that satisfy (\ref{eq:cc_v})~and~(\ref{eq:cc_r}) for class $k^*(\sigma_i)$.
\end{remark}

In order for a case, $\{\mathcal{N}^t_1(\sigma_i),\ldots,\mathcal{N}^t_K(\sigma_i)\}$, to result in subregion $\sigma_i$ being consistently classified incorrectly, (\ref{eq:cc_v}) and (\ref{eq:cc_r}) must be satisfied for an incorrect class $j$, where $j\in\{1,\ldots,K\}$ and $j\neq k^*(\sigma_i)$.
\begin{remark}
    The probability of consistently classifying incorrectly is 
    \begin{equation}\label{eq:prob_incorr}
    \begin{split}
        P(\sigma_i\text{ is consistently}&\text{ classified incorrectly}) \\
        &= \sum_{\substack{j \in \{1,\ldots,K\} \\ j \neq k^*(\sigma_i)}}
        ~~\sum_{t:\{\mathcal{N}^t_1(\sigma_i),\ldots,\mathcal{N}^t_K(\sigma_i)\}\in\mathcal{A}(j,\sigma_i)}~~\prod_{k=1}^K ~~\prod_{n\in\mathcal{N}^t_k(\sigma_i)}p_n^k(\sigma_i)
    \end{split}
    \end{equation}
    where $\mathcal{A}(j,\sigma_i)$ is the set of cases, $\{\mathcal{N}^t_1(\sigma_i),\ldots,\mathcal{N}^t_K(\sigma_i)\}$, $t\in\{1,\ldots,K^N\}$, that satisfy (\ref{eq:cc_v})~and~(\ref{eq:cc_r}) for class $j\in\{1,\ldots,K\}$, $j\neq k^*(\sigma_i)$.
\end{remark}

In order for a case to be inconsistent, there must be no $k$ that satisfies (\ref{eq:cc_v}) and (\ref{eq:cc_r}).
\begin{remark}
    The probability of inconsistent classification is 
    \begin{align}\label{eq:prob_incon}
        P(\sigma_i\text{ is inconsistently classified})=1&-P(\sigma_i\text{ is consistently}\text{ classified correctly as $k^*(\sigma_i)$}) \nonumber \\ 
        &-P(\sigma_i\text{ is consistently}\text{ classified incorrectly}).
    \end{align}
\end{remark}

In Section~\ref{s:Empirical_Results}, we implement S-BOMM using PBnB as the sampling and classification method. This choice is advantageous because the theoretic guarantees provided in~\cite{PBnB_Chapter} provides a useful bound, specifically,
\begin{equation}\label{eq:PBnB_maint}
    P(Y_n(\sigma_i)=k_n^{\text{T}}(\sigma_i))\ge (1-\alpha)^4.
\end{equation}
With this information, we can bound the probabilities in (\ref{eq:prob_corr}) and (\ref{eq:prob_incorr}).
\begin{proposition}
    Under PBnB sampling and classification methodology, a bound on the probability of consistently classifying correctly is
    \begin{equation*}\label{eq:prop_cc_corr}
    \begin{split}
        P(\sigma_i&\text{ is consistently classified correctly as $k^*(\sigma_i)$}) \\
        & \ge \sum_{t:\{\mathcal{N}^t_1(\sigma_i),\ldots,\mathcal{N}^t_K(\sigma_i)\}\in\mathcal{A}(k^*(\sigma_i),\sigma_i)}
        ~~\prod_{k=1}^K 
        ~~\prod_{\substack{n\in\mathcal{N}^t_k(\sigma_i)\\ \text{and }n\in\mathcal{B}(k,\sigma_i)}}(1-\alpha)^4
        \prod_{\substack{n\in\mathcal{N}^t_k(\sigma_i)\\ \text{and }n\notin\mathcal{B}(k,\sigma_i)}}p_n^k(\sigma_i)
    \end{split}
    \end{equation*}
    where $\mathcal{B}(k,\sigma_i)$ is the set of models, $n\subseteq\{1,\ldots,N\}$, where $k^\text{T}_n(\sigma_i)=k$.
\end{proposition}
\begin{proposition}
    Under PBnB sampling and classification methodology, a bound on the probability of consistently classifying incorrectly is
        \begin{equation*}\label{eq:prop_cc_incorr}
    \begin{split}
        P(\sigma_i&\text{ is consistently classified incorrectly}) \\
        & \ge \sum_{\substack{j \in \{1,\ldots,K\} \\ j \neq k^*(\sigma_i)}}
        ~\sum_{t:\{\mathcal{N}^t_1(\sigma_i),\ldots,\mathcal{N}^t_K(\sigma_i)\}\in\mathcal{A}(j,\sigma_i)}
        ~~\prod_{k=1}^K 
        ~~\prod_{\substack{n\in\mathcal{N}^t_k(\sigma_i)\\ \text{and }n\in\mathcal{B}(k,\sigma_i)}}(1-\alpha)^4
        \prod_{\substack{n\in\mathcal{N}^t_k(\sigma_i)\\ \text{and }n\notin\mathcal{B}(k,\sigma_i)}}p_n^k(\sigma_i)
    \end{split}
    \end{equation*}
    where $\mathcal{B}(k,\sigma_i)$ is the set of models, $n\subseteq\{1,\ldots,N\}$, where $k^\text{T}_n(\sigma_i)=k$.
\end{proposition}

To further evaluate these expressions, we construct an example and numerically evaluate the probabilities in (\ref{eq:prob_corr}), (\ref{eq:prob_incorr}), and (\ref{eq:prob_incon}) for specific values of $v$ and $r$, and in doing so, illustrate the impact of $v$ and $r$.

Consider $K=3$ classes and $N=3$ models. This choice provides sufficient complexity for illustration while remaining combinatorially tractable. With $K=3$ classes and $N=3$ models, there are $27$ cases to consider.

Let $k^*(\sigma_i)=1$ be the correct consistent class of subregion $\sigma_i$. 
There are two relevant scenarios we consider that would result in $k^*(\sigma_i)=1$.
The first is when all three models agree, that is, $k_n^\text{T}(\sigma_i)=1$ for all $n\in\{1,2,3\}$.
The second is when two of the three models agree, that is, $k_n^\text{T}(\sigma_i)=1$ for two $n\in\{1,2\}$ and $k_n^\text{T}(\sigma_i)\neq 1$ for $n=3$.
These two scenarios become relevant for purposes of  estimating  $p_n^k(\sigma_i)$.

We let $p_1^{3}(\sigma_i)=p_2^{3}(\sigma_i)=p_3^{3}(\sigma_i)=0.15$ and we let $\alpha = 0.1$.
For the scenario in which $k_1^T(\sigma_i)=1$, $k_2^T(\sigma_i)=1$, and $k_3^T(\sigma_i)=1$, we have $k_1^1(\sigma_i)=k_2^1(\sigma_i)=k_3^1(\sigma_i)=(1-\alpha)^4=0.6561$, which implies $p_1^{2}(\sigma_i)=p_2^{2}(\sigma_i)=p_3^{2}(\sigma_i)=1-(1-\alpha)^4-0.15=0.1939$.
For the scenario in which $k_1^T(\sigma_i)=1$, $k_2^T(\sigma_i)=1$, and $k_3^T(\sigma_i)=2$, we have $k_1^1(\sigma_i)=k_2^1(\sigma_i)=k_3^2(\sigma_i)=(1-\alpha)^4=0.6561$, which implies ${p_1^{2}(\sigma_i)=p_2^{2}(\sigma_i)=p_3^{1}(\sigma_i)=1-(1-\alpha)^4-0.15=0.1939}$.

Given values for $p_n^k(\sigma_i)$ for each scenario as defined above,
the consistency scores $C^t_k(\sigma_i)$ can be calculated.  Then, using $C^t_k(\sigma_i)$ and parameters $v$ and $r$, we determine which cases satisfy (\ref{eq:cc_v}) and (\ref{eq:cc_r}) and evaluate the probability expressions in (\ref{eq:prob_corr}), (\ref{eq:prob_incorr}), and (\ref{eq:prob_incon}).

\begin{figure}
\begin{subfigure}{.48\linewidth}
    \centering
    \includegraphics[width=\linewidth]{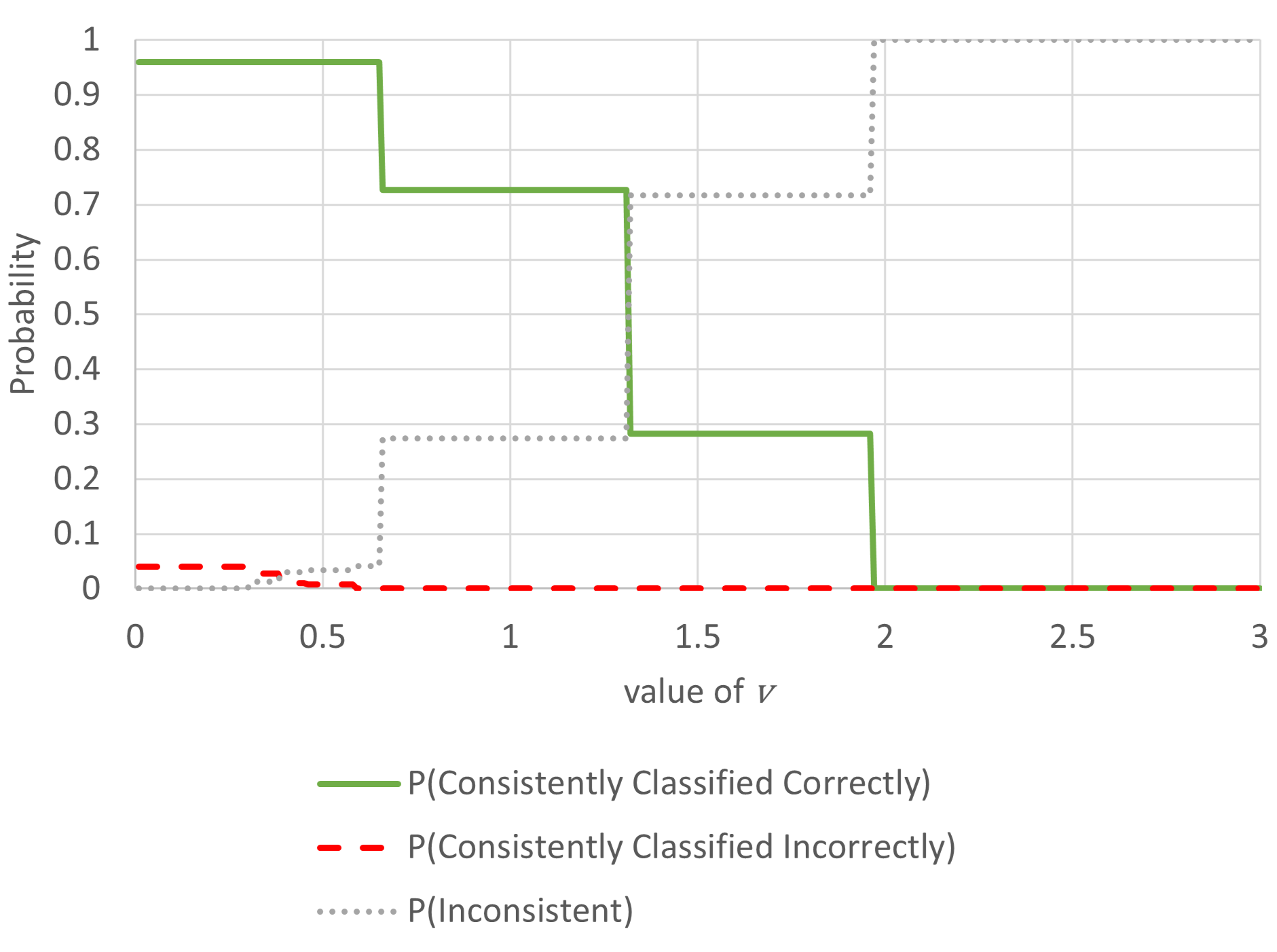}
    \caption{Varying $v$ with fixed $r=0.01$ when ${k_1^T(\sigma_i)=1}$, $k_2^T(\sigma_i)=1$, and $k_3^T(\sigma_i)=1$}
    \label{fig:results_3n_v}
\end{subfigure}
\hfill
\begin{subfigure}{.48\linewidth}
    \centering
    \includegraphics[width=\linewidth]{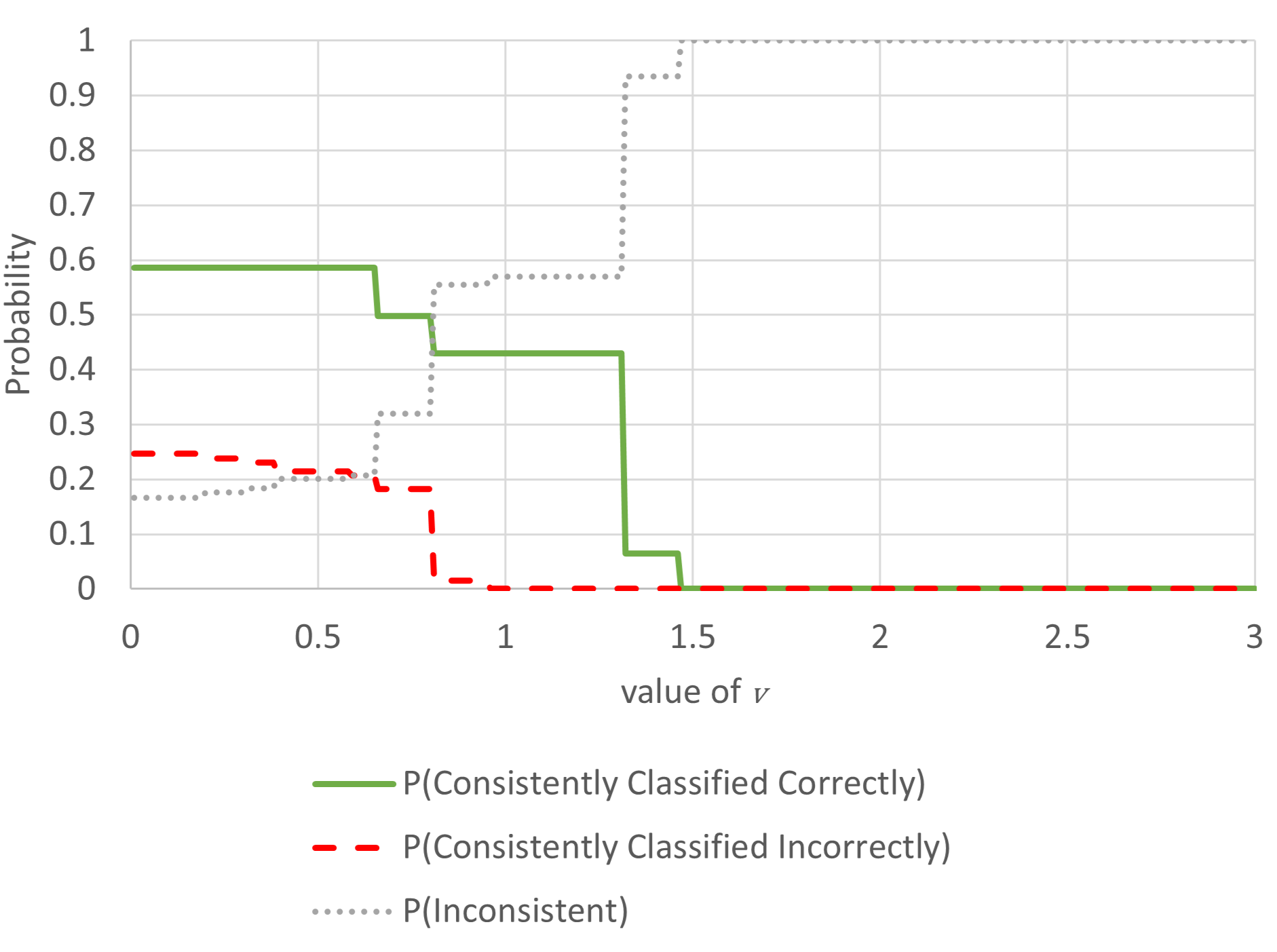}
    \caption{Varying $v$ with fixed $r=0.01$ when ${k_1^T(\sigma_i)=1}$, $k_2^T(\sigma_i)=1$, and $k_3^T(\sigma_i)=2$}
    \label{fig:results_2n_v}    
\end{subfigure}
\\
\begin{subfigure}{.48\linewidth}
    \centering
    \includegraphics[width=\linewidth]{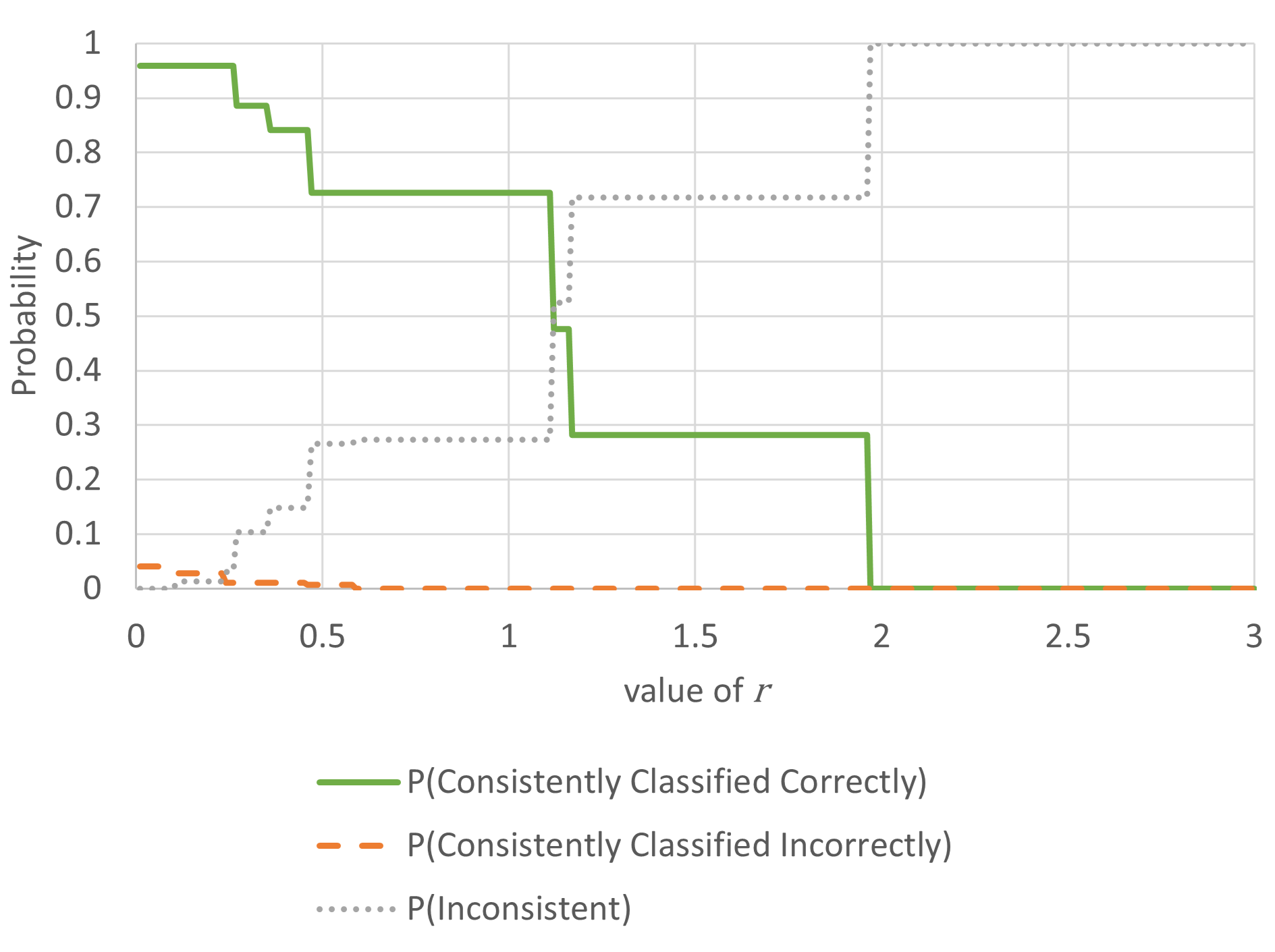}
    \caption{Varying $r$ with fixed $v=0.01$ when ${k_1^T(\sigma_i)=1}$, $k_2^T(\sigma_i)=1$, and $k_3^T(\sigma_i)=1$}
    \label{fig:results_3n_r}
\end{subfigure}
\hfill
\begin{subfigure}{.48\linewidth}
    \centering
    \includegraphics[width=\linewidth]{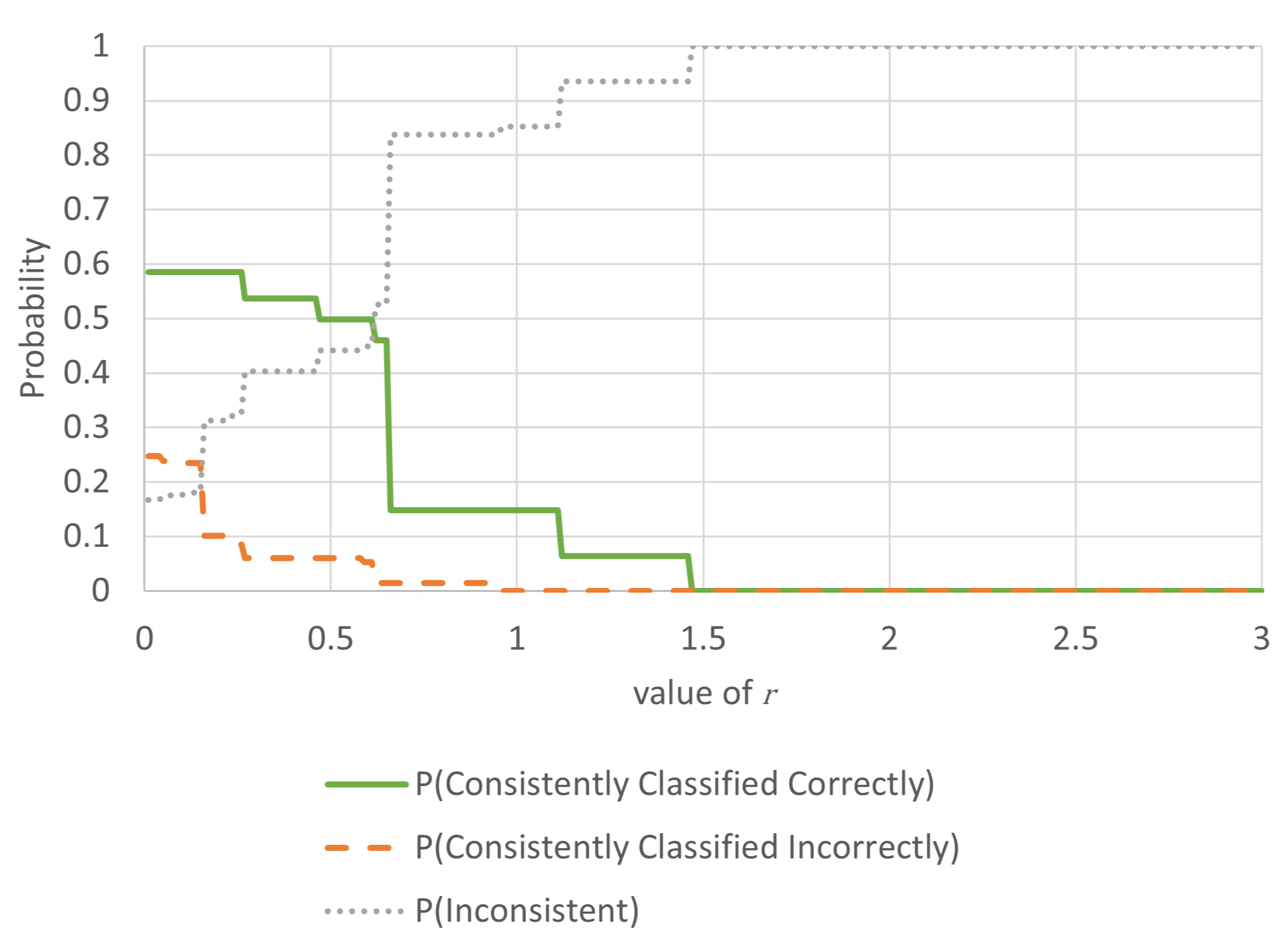}
    \caption{Varying $r$ with fixed $v=0.01$ when ${k_1^T(\sigma_i)=1}$, $k_2^T(\sigma_i)=1$, and $k_3^T(\sigma_i)=2$}
    \label{fig:results_2n_r}    
\end{subfigure}
    \centering
    \caption{Impact of $v$ and $r$ on the probability of consistently classifying correctly as in (\ref{eq:prob_corr}), consistently classifying incorrectly as in (\ref{eq:prob_incorr}), and inconsistently classifying as in (\ref{eq:prob_incon})}
    \label{fig:results_all_four}
\end{figure}

The probability trajectories of consistently classifying correctly as in (\ref{eq:prob_corr}), consistently classifying incorrectly as in (\ref{eq:prob_incorr}), and inconsistently classifying as in (\ref{eq:prob_incon}) are provided in Figure~\ref{fig:results_all_four} as functions of parameters $v$ and $r$. 
More specifically, Figure~\ref{fig:results_3n_v} shows the probabilities for different values of $v$ holding $r$ constant under the scenario where all three models agree, 
Figure~\ref{fig:results_3n_r} shows the probabilities for different values of $r$ holding $v$ constant under the scenario where all three models agree.
For the second scenario where two of the three models agree,
Figure~\ref{fig:results_2n_v} shows the probabilities for different values of $v$ holding $r$ constant
and Figure~\ref{fig:results_2n_r} shows the probabilities 
for different values of $r$ holding $v$ constant. 

Figure~\ref{fig:results_all_four} highlights the impact of $v$ and $r$ on the performance of the consistency methodology.  The lowest values of $v$ and $r$ elicit the maximum probability of consistently classifying correctly, but also results in the largest probability of consistently classifying incorrectly. The probability of consistently classifying incorrectly can be reduced by increasing the value of $v$ and/or $r$ at the cost of increased probability of inconsistent classification and reduced probability of consistently classifying correctly. If $v$ or $r$ are too high (e.g., $v\ge2$ or $r\ge2$) it becomes impossible to satisfy condition~(\ref{eq:cc_v})~or~(\ref{eq:cc_r}), respectively, and will thus always result in inconsistent classification.

\section{Empirical Results}\label{s:Empirical_Results}
To demonstrate the capabilities of the S-BOMM methodology beyond the assumptions required for the analysis in Section~\ref{s:analysis}, 
we implement S-BOMM with a sampling and classification method from PBnB\citep{PBnB_Chapter}. We apply this implementation to two examples, presented in Sections~\ref{s:example1}~and~\ref{s:example2}, and then discuss the results in Section~\ref{s:discussion}.

We selected PBnB for the implementation of branching, sampling, and classifying for this paper because its set-based nature is well suited to S-BOMM and there are theoretical guarantees that are helpful in estimating $p_n^k(\sigma_i)$.

The branching scheme and the sampling and classification method in PBnB is detailed in Section~6.3 of~\cite{PBnB_Chapter}. The PBnB parameters are set to $\delta = 0.2$, $\alpha = 0.1$, $\epsilon = 0.4$, $B=2$, and $c=20$.

We estimate $p_n^k(\sigma_i)$ using the confidence intervals developed in~\cite{PBnB_Chapter}. Specifically, the confidence interval of the target region is calculated via (6.7)-(6.9) in \cite{PBnB_Chapter}.
In our implementation, if the subregion is classified as contained in the target region, then we set the lower bound of the confidence interval to  the largest function value sampled in the subregion and let $(1-\alpha^\prime)$ be the confidence level  that results in this confidence interval lower bound.
In the numerical examples, the value of $\alpha^\prime$ is less than or equal to the input $\alpha=0.1$, reflecting that the required confidence has been met and the actual confidence is greater.
Similarly, if the subregion is classified as outside the target region, then we set the upper bound of the confidence interval to the smallest function value sampled in the subregion and and let $(1-\alpha^\prime)$ be the confidence level that results in this confidence interval upper bound. We estimate $p_n^k(\sigma_i)=1-\alpha^\prime$.

\subsection{\emph{Example 1: 3 Models and 3 Classes}}\label{s:example1}
We present an example with three models ($N=3$) and three classes ($K=3$). 
The target regions are the 20th percentiles of minimal output values on each model.
The first class indicates regions that reside within the target region. The second class indicates regions that reside entirely outside the target region. The third class indicates regions that cannot be classified under the first two classes. We set $v=1$ and $r=1$. 
The stopping criteria is that at least $10\%$ of the decision space has been consistently classified as within the target region. 

\begin{figure}
    \begin{subfigure}{.32\textwidth}
    \centering
    \includegraphics[width=0.97\linewidth]{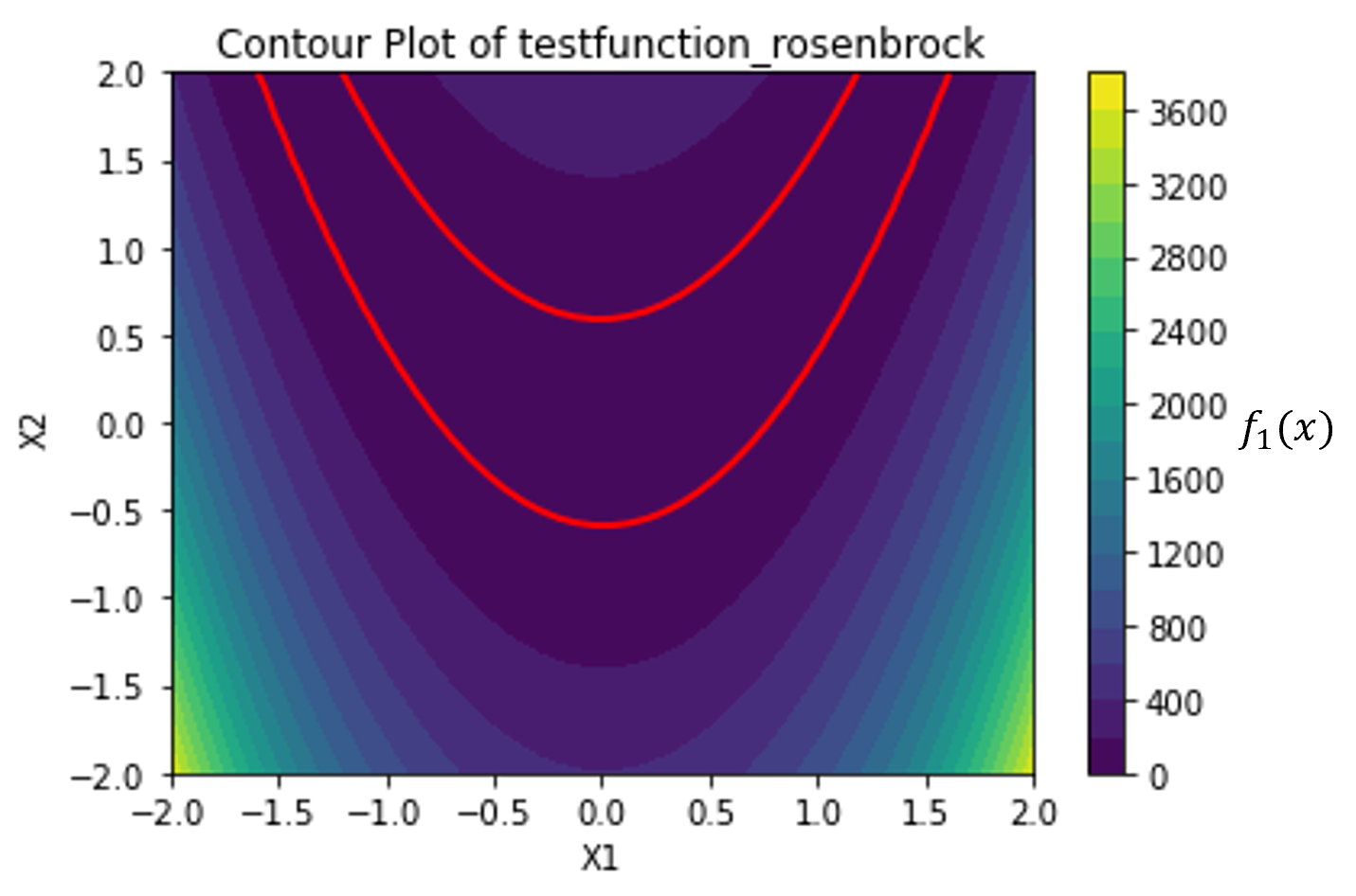}
    \caption{Model 1 as given by (\ref{eq:mod1})}
    \label{fig:mod1}
    \end{subfigure}
    \hfill
    \begin{subfigure}{.32\textwidth}
    \centering
    \includegraphics[width=0.97\linewidth]{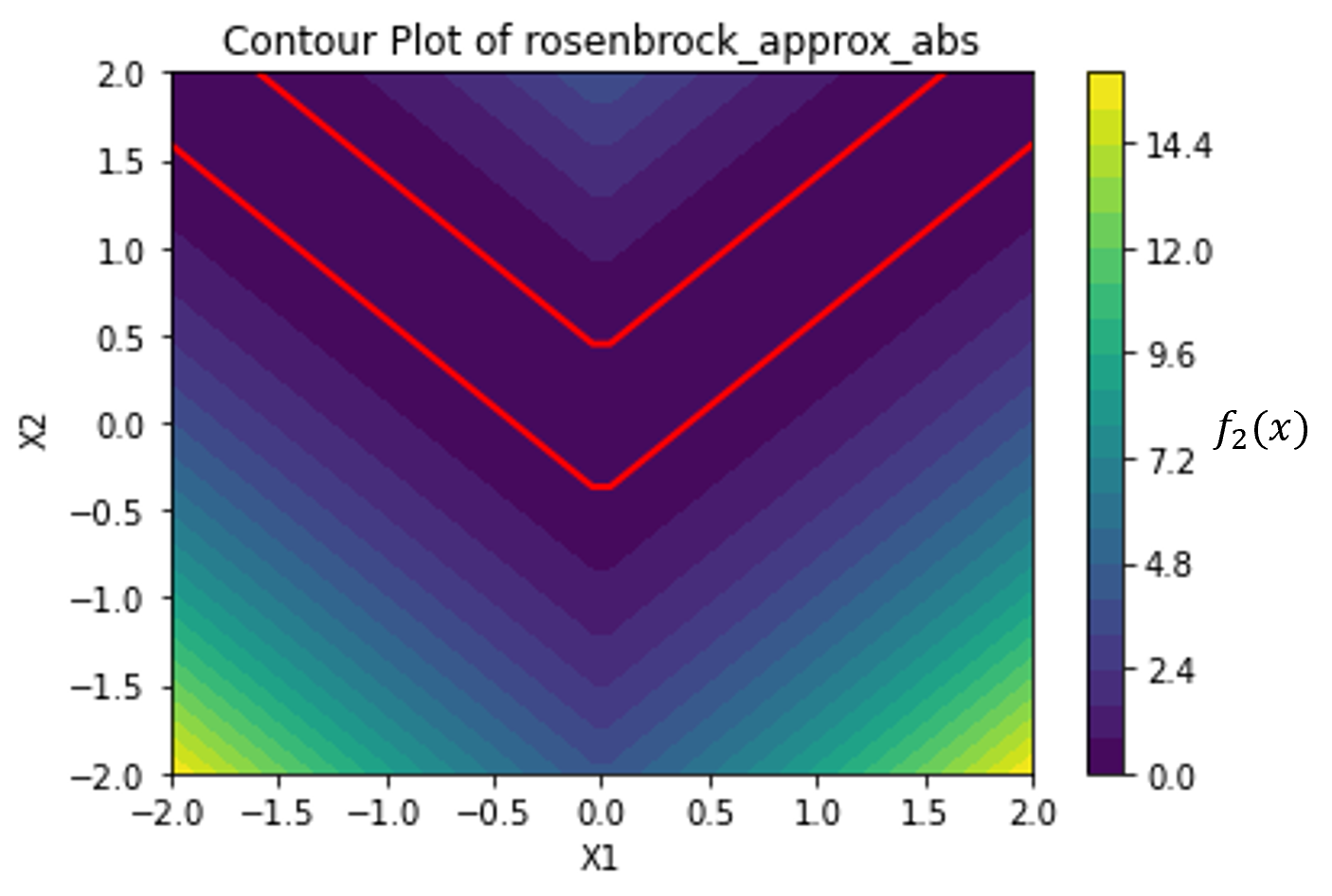}
    \caption{Model 2 as given by (\ref{eq:mod2})}
    \label{fig:mod2}
    \end{subfigure}
    \hfill
    \begin{subfigure}{.32\textwidth}
    \centering
    \includegraphics[width=0.97\linewidth]{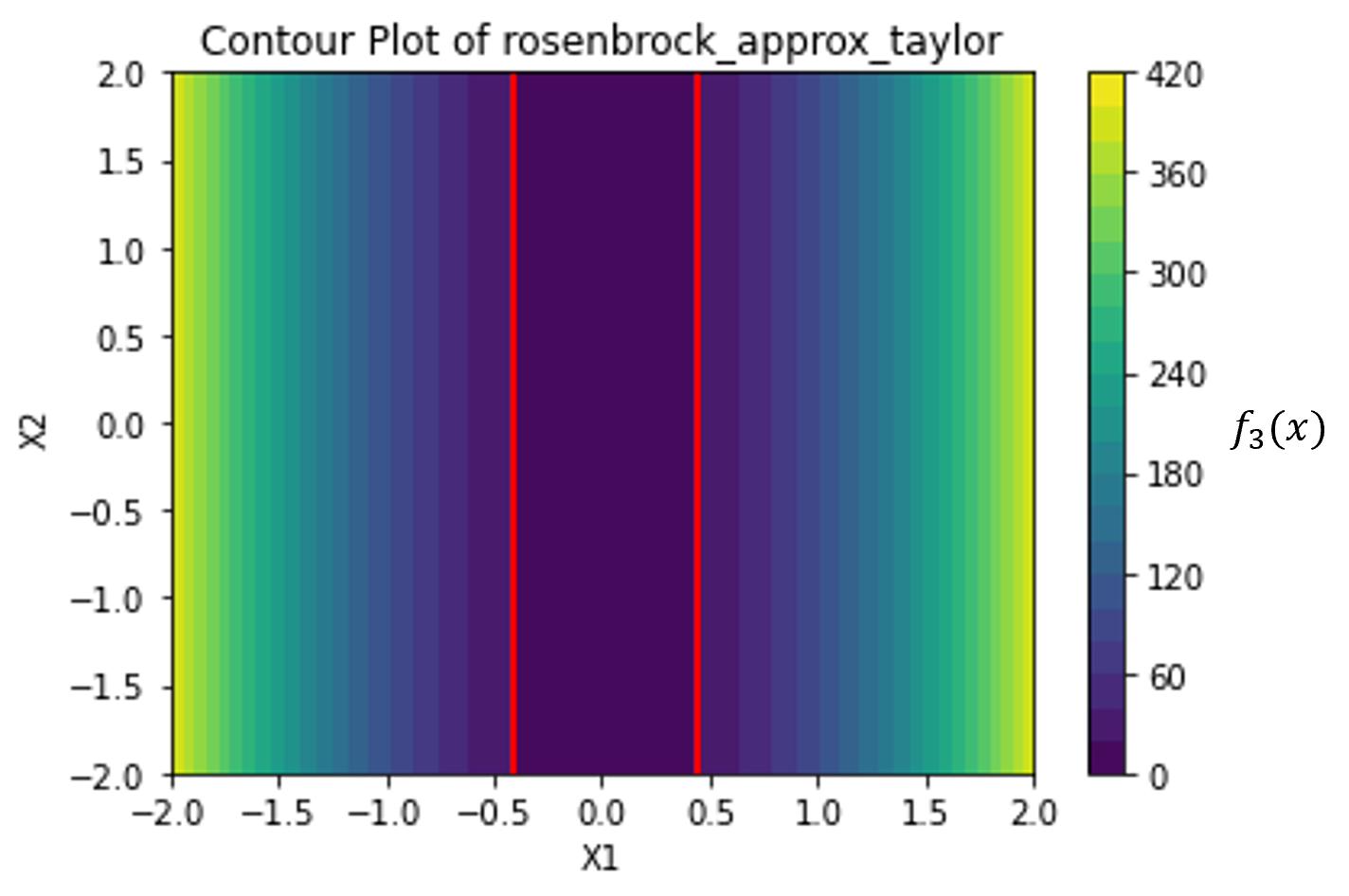}
    \caption{Model 3 as given by (\ref{eq:mod3})}
    \label{fig:mod3}
    \end{subfigure}
    \caption{Heatmap of the objective functions of the three models (\ref{eq:mod1}), (\ref{eq:mod2}), (\ref{eq:mod3}) in Example~1 with the red contour indicating the 20th percentile }
    \label{fig:models}
\end{figure}

\begin{figure}
\begin{subfigure}[t]{.48\textwidth}
    \centering
    \includegraphics[width=0.92\linewidth]{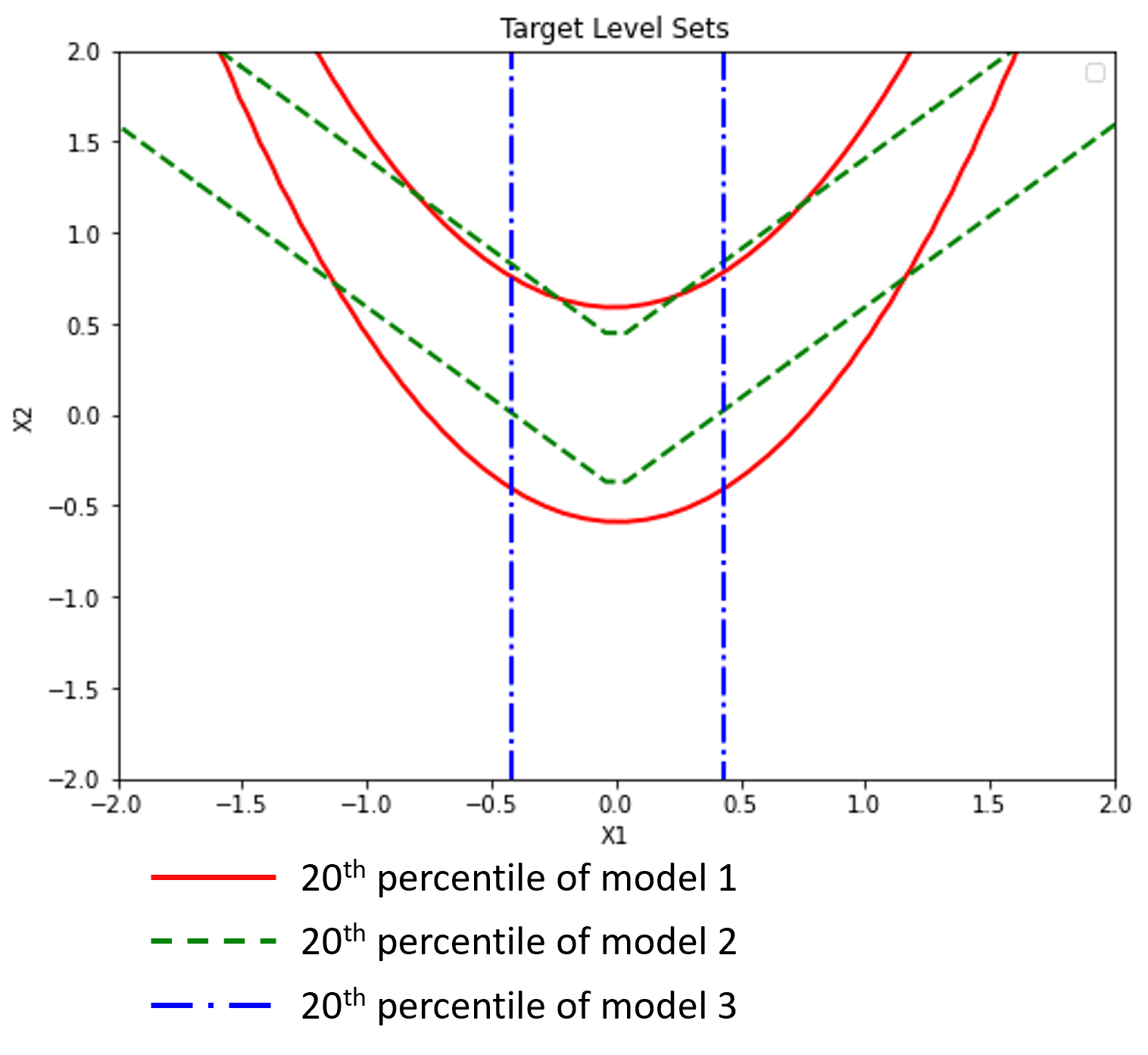}
    \caption{Target regions of each model in Example 1.}
    \label{fig:levelsets}
\end{subfigure}
\hfill
\begin{subfigure}[t]{.48\textwidth}
    \centering
    \includegraphics[width=0.99\linewidth]{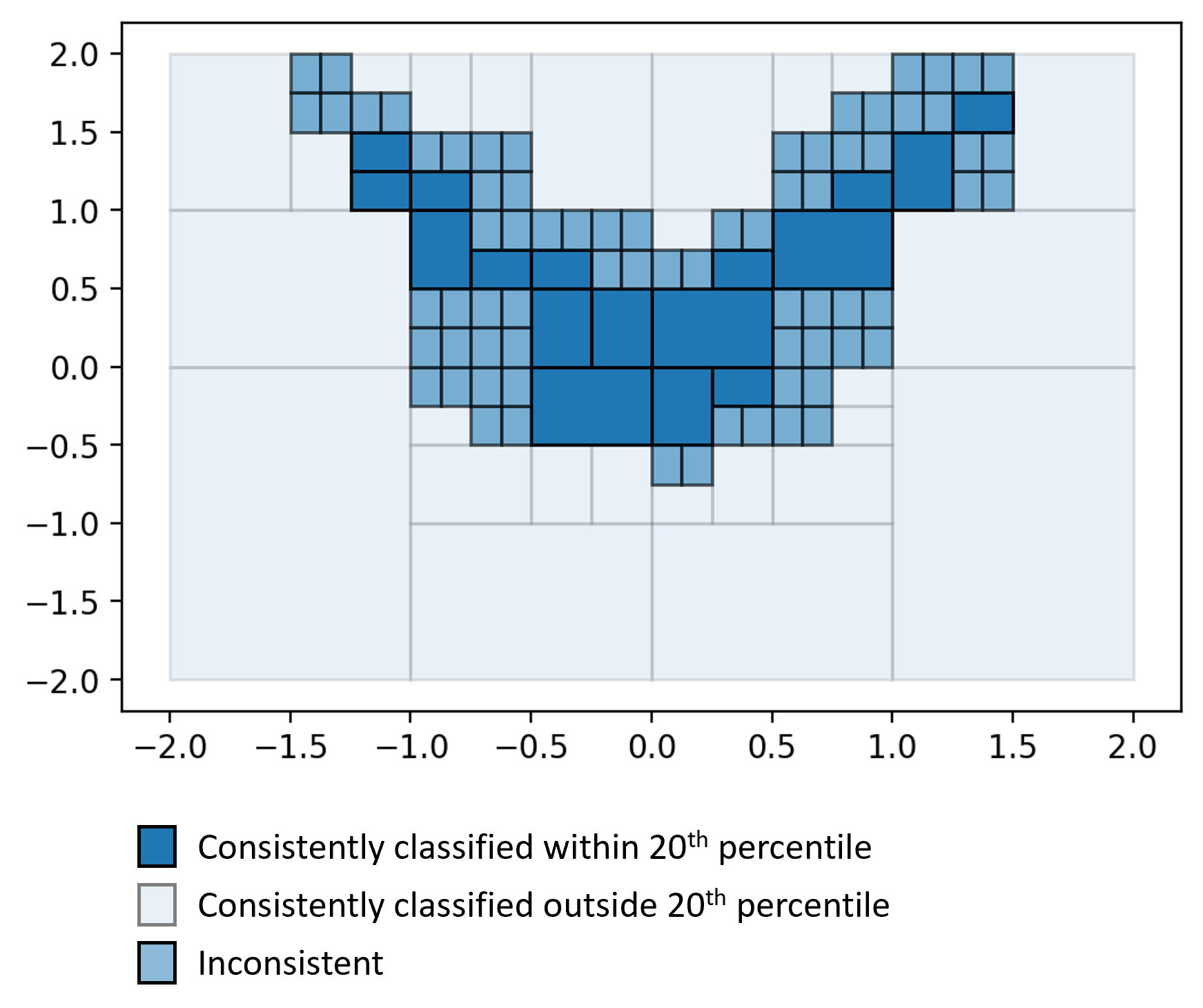}
    \caption{Example 1 results after termination at 10 iterations.}
    \label{fig:empiricalresults}
\end{subfigure}
\caption{Example 1 target regions (a) and solution set (b).}
\end{figure}

The first model is a two-dimensional Rosenbrock function, defined as
\begin{equation}\label{eq:mod1}
    f_1(x_1,x_2)=(1-x_1)^2+100(x_2-x_1^2)^2.
\end{equation}
The second model is a two-dimensional quadratic absolute value function defined as
\begin{equation}\label{eq:mod2}
    f_2(x_1,x_2)=(|x_1|-x_2)^2.
\end{equation}
The third model is a second order Taylor series approximation of the Rosenbrock function at the origin, defined as
\begin{equation}\label{eq:mod3}
    f_3(x_1,x_2)=1-2x_1+100x_1^2.
\end{equation}
These models are illustrated in Figure~\ref{fig:models} with the 20th percentile outlined in red. All three models are defined on $-2\le x_i \le 2$ for $i=1,2$. Figure~\ref{fig:levelsets} illustrates the target region for each model.

After 10 iterations, S-BOMM terminated by successfully consistently classifying approximately $10\%$ of the input domain as within the 20th percentile. These results are shown in Figure~\ref{fig:empiricalresults}. Notice that there are large regions near $x_1=0$ and $x_2=0$ that are consistently classified within the 20th percentile, which reflects that all three models agree that regions near $x_1=0$ and $x_2=0$ are in the target region. There are several consistently classified regions to the upper left and right of the origin because Models 1 and 2 both classified these regions as within the 20th percentile, even though these regions are outside of the target region according to Model 3. 

The large subregion in the bottom left, from $(-2.0,-2.0)$ to $(-1.0,0.0)$, was consistently classified as outside the 20th percentile on the 3rd iteration. The consistency score corresponding to the class inside the 20th percentile, $C^t_1(\sigma_i)$ where $\sigma_i$ is the subregion  $-2.0\le x_1\le -1.0$ and $-2.0\le x_2\le 0.0$, is $C^t_1(\sigma_i)=0$ and the consistency score corresponding to the class outside the 20th percentile, $C^t_2(\sigma_i)$ on the same subregion $\sigma_i$, $-2.0\le x_1\le -1.0$ and $-2.0\le x_2\le 0.0$, is $C^t_2(\sigma_i)=2.97$. Such a high consistency score corresponding to the class outside the 20th percentile means that all three models agreed with a high probability that the subregion is outside the 20th percentile and therefore it was able to be consistently classified on only the 3rd iteration. This saved computation in subsequent iterations.

On the other hand, consider the subregion from $(1.375,1.75)$ to $(1.5,2.0)$ near the upper right hand corner. This subregion was inconsistently classified after all 10 iterations. Model 1 and Model 2 classified as within the 20th percentile with relatively low probability, leading to a consistency score of $C^t_1(\sigma_i)=0.992$. Model 3 classified as outside the 20th percentile with very high probability, leading to a consistency score of $C^t_2(\sigma_i)=0.99$. These scores failed to satisfy (\ref{eq:cc_v}) and (\ref{eq:cc_r}), so this region is inconsistently classified.

\subsection{\emph{Example 2: 4 Models and 3 Classes}}\label{s:example2}
We now present an example with four models ($N=4$) and three classes ($K=3$). 
The target regions are the 20th percentiles of minimal output values for each model.
As before, the first class indicates regions that reside within the target region. The second class indicates regions that reside entirely outside the target region. The third class indicates regions that cannot be classified under the first two classes. We set $v=1$ and $r=0.75$. A lower $r$ parameter value was chosen for this example to accommodate greater discrepancy between models. 
The stopping criteria is once again that at least $10\%$ of the decision space has been consistently classified as within the target region.

\begin{figure}
    \begin{subfigure}{.47\textwidth}
    \centering
    \includegraphics[width=0.99\linewidth]{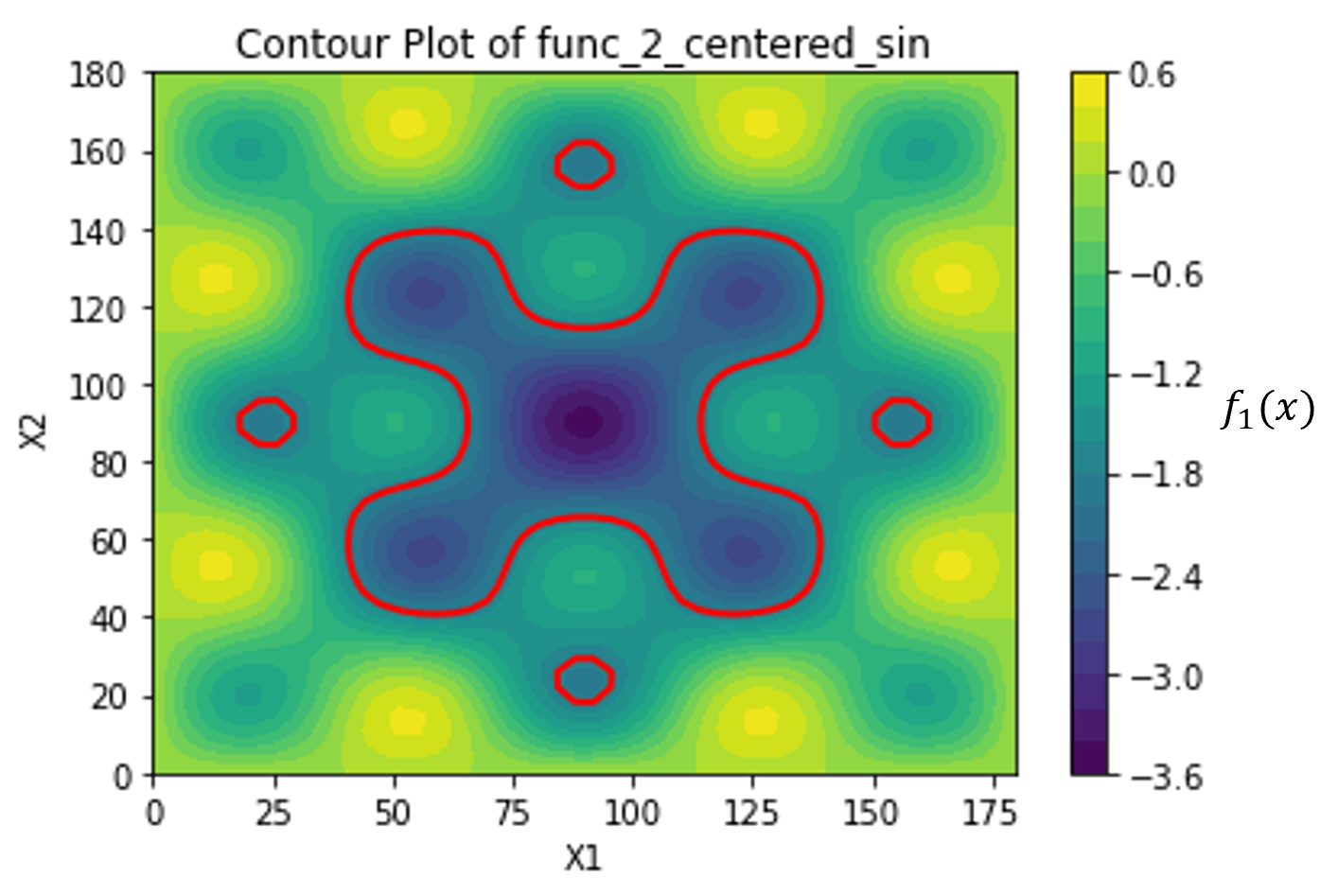}
    \caption{Model 1 as given by (\ref{eq:ex2mod1})}
    \label{fig:ex2mod1}
    \end{subfigure}
    \hfill
    \begin{subfigure}{.47\textwidth}
    \centering
    \includegraphics[width=0.99\linewidth]{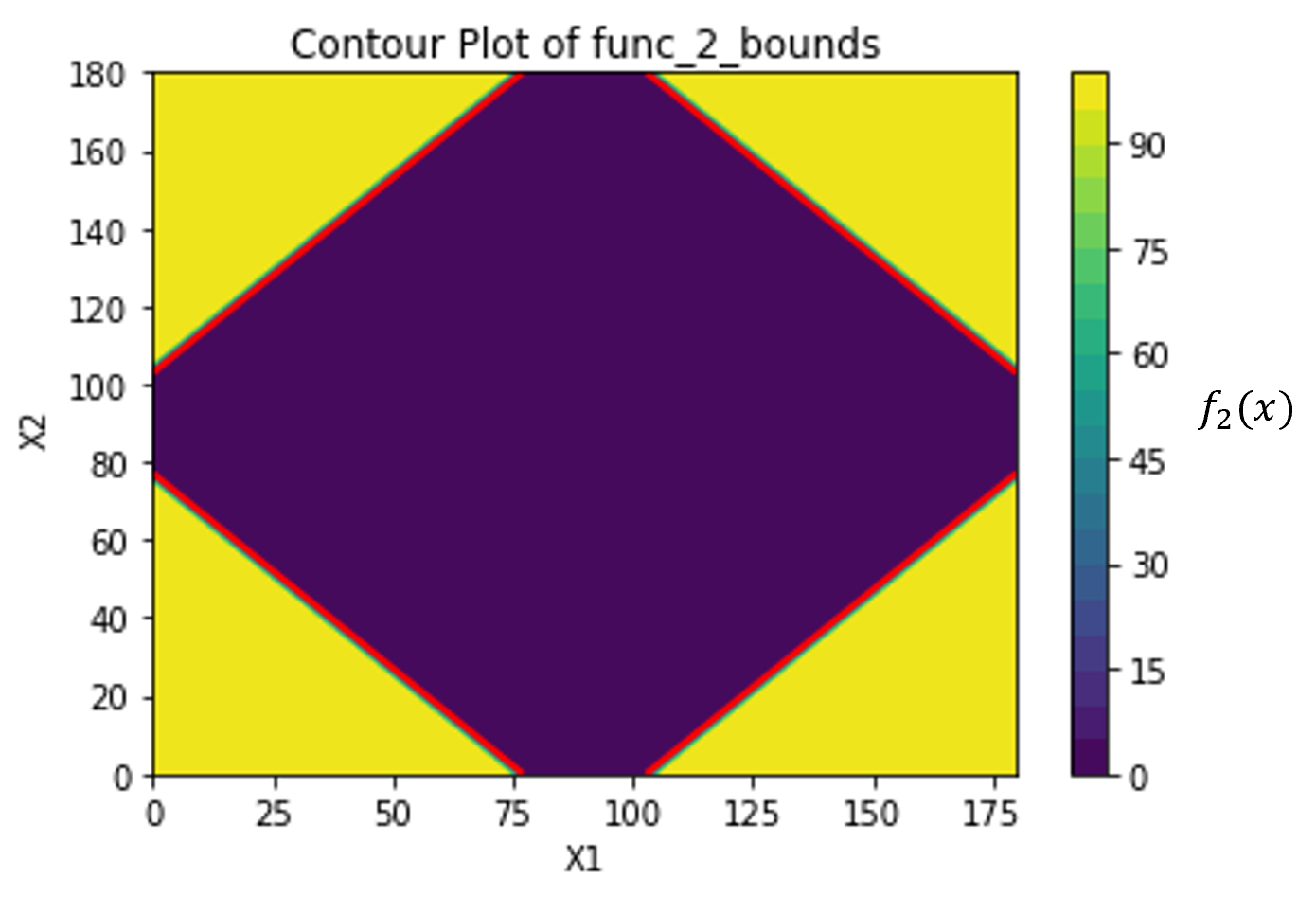}
    \caption{Model 2 as given by (\ref{eq:ex2mod2})}
    \label{fig:3x2mod2}
    \end{subfigure}
    \\
    \begin{subfigure}{.47\textwidth}
    \centering
    \includegraphics[width=0.99\linewidth]{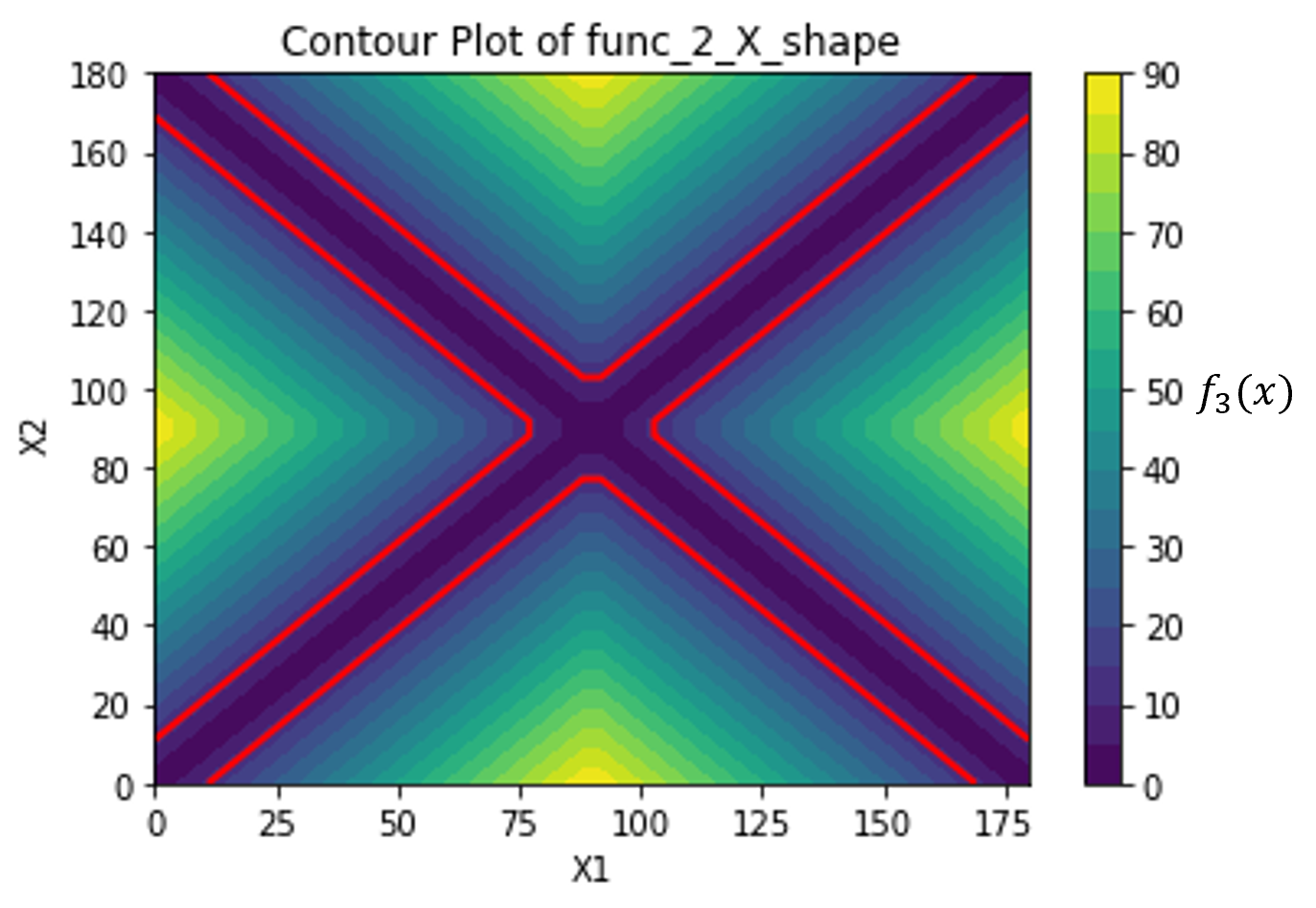}
    \caption{Model 3 as given by (\ref{eq:ex2mod3})}
    \label{fig:ex2mod3}
    \end{subfigure}
    \hfill
    \begin{subfigure}{.47\textwidth}
    \centering
    \includegraphics[width=0.99\linewidth]{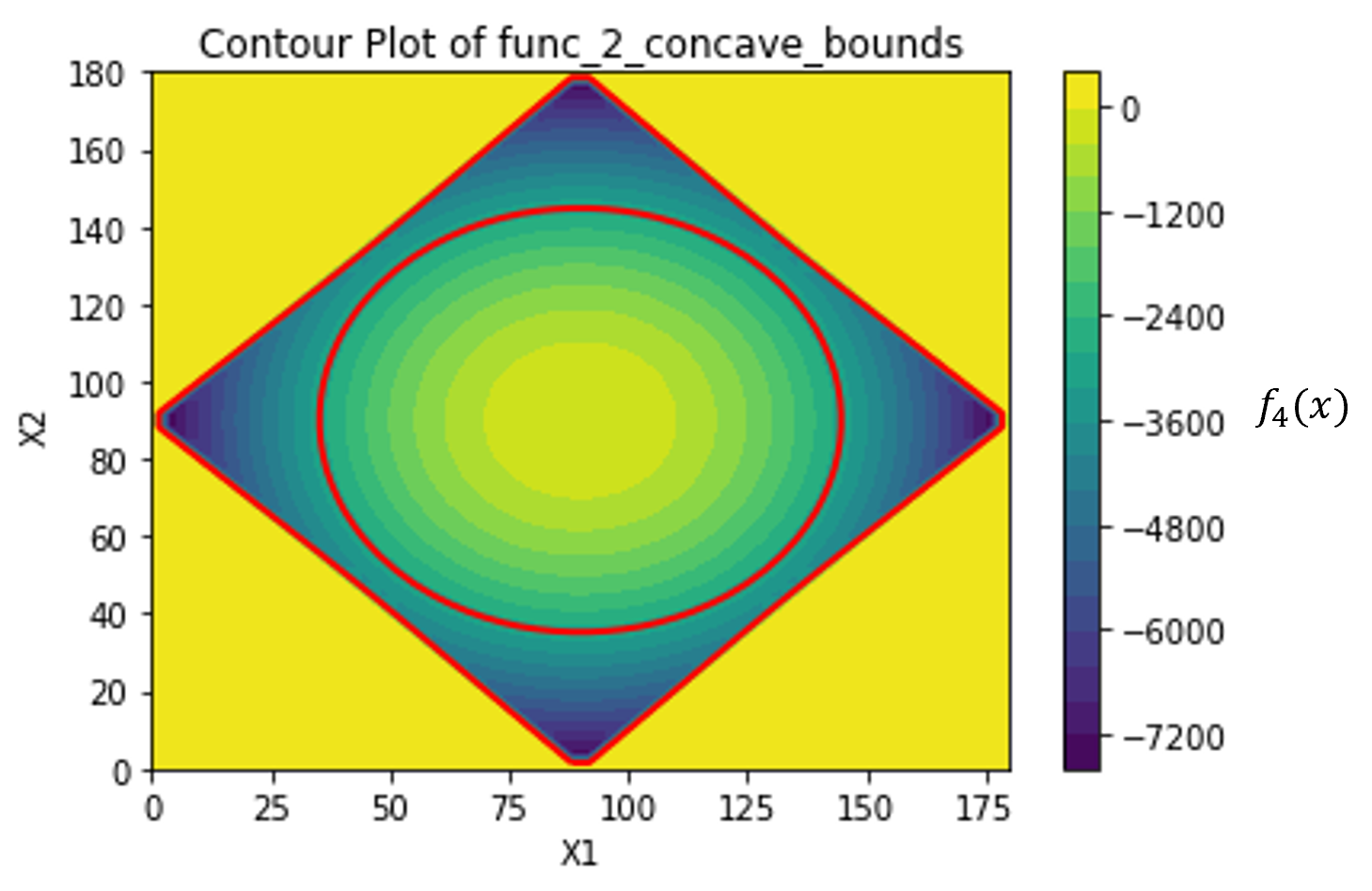}
    \caption{Model 4 as given by (\ref{eq:ex2mod4})}
    \label{fig:ex2mod4}
    \end{subfigure}
    \caption{Heatmap of the objective function of the four models (\ref{eq:ex2mod1}), (\ref{eq:ex2mod2}), (\ref{eq:ex2mod3}), (\ref{eq:ex2mod4}) in Example 2 with the red contour indicating the 20th percentile.}
    \label{fig:models2}
\end{figure}

The first model is a two-dimensional centered sinusoidal function, defined as
\begin{equation}\label{eq:ex2mod1}
    f_1(x_1,x_2)=-2.5\sin\left(\pi\frac{x_1}{180}\right)\sin\left(\pi\frac{x_2}{180}\right)-\sin\left(\pi\frac{x_1}{36}\right)\sin\left(\pi\frac{x_2}{36}\right).
\end{equation}
The second model is a piecewise constant function, 
defined as
\begin{equation}\label{eq:ex2mod2}
    f_2(x_1,x_2)=\left\{\begin{array}{cl}
        100 & \text{if }x_1+x_2<75 \\
        100 & \text{if }x_1-x_2<105 \\
        100 & \text{if }-x_1+x_2<105 \\
        100 & \text{if }x_1+x_2<285 \\
        0 & \text{otherwise}. \\
    \end{array}\right.
\end{equation}
The third model is an absolute value function defined as 
\begin{equation}\label{eq:ex2mod3}
    f_3(x_1,x_2)=|(|x_1-90|)-(|x_2-90|)|.
\end{equation}
The fourth model is a piecewise quadratic function, defined as
\begin{equation}\label{eq:ex2mod4}
    f_2(x_1,x_2)=\left\{\begin{array}{cl}
        100 & \text{if }x_1+x_2<90 \\
        100 & \text{if }x_1-x_2<90 \\
        100 & \text{if }-x_1+x_2<90 \\
        100 & \text{if }x_1+x_2<270 \\
        -(x_1-90)^2-(x_2-90)^2 & \text{otherwise.} \\
    \end{array}\right.
\end{equation}
These models are illustrated in Figure~\ref{fig:models2} with the 20th percentile outlined in red. All four models are defined on $0\le x_i \le 180$ for $i=1,2$. Notice that the target regions for Models~1,~3,~and~4 are non-convex, and the  target region for Model~1 is disconnected, making this example challenging for many optimization methods.

\begin{figure}
\begin{subfigure}{.43\textwidth}
    \centering
    \includegraphics[width=0.99\linewidth]{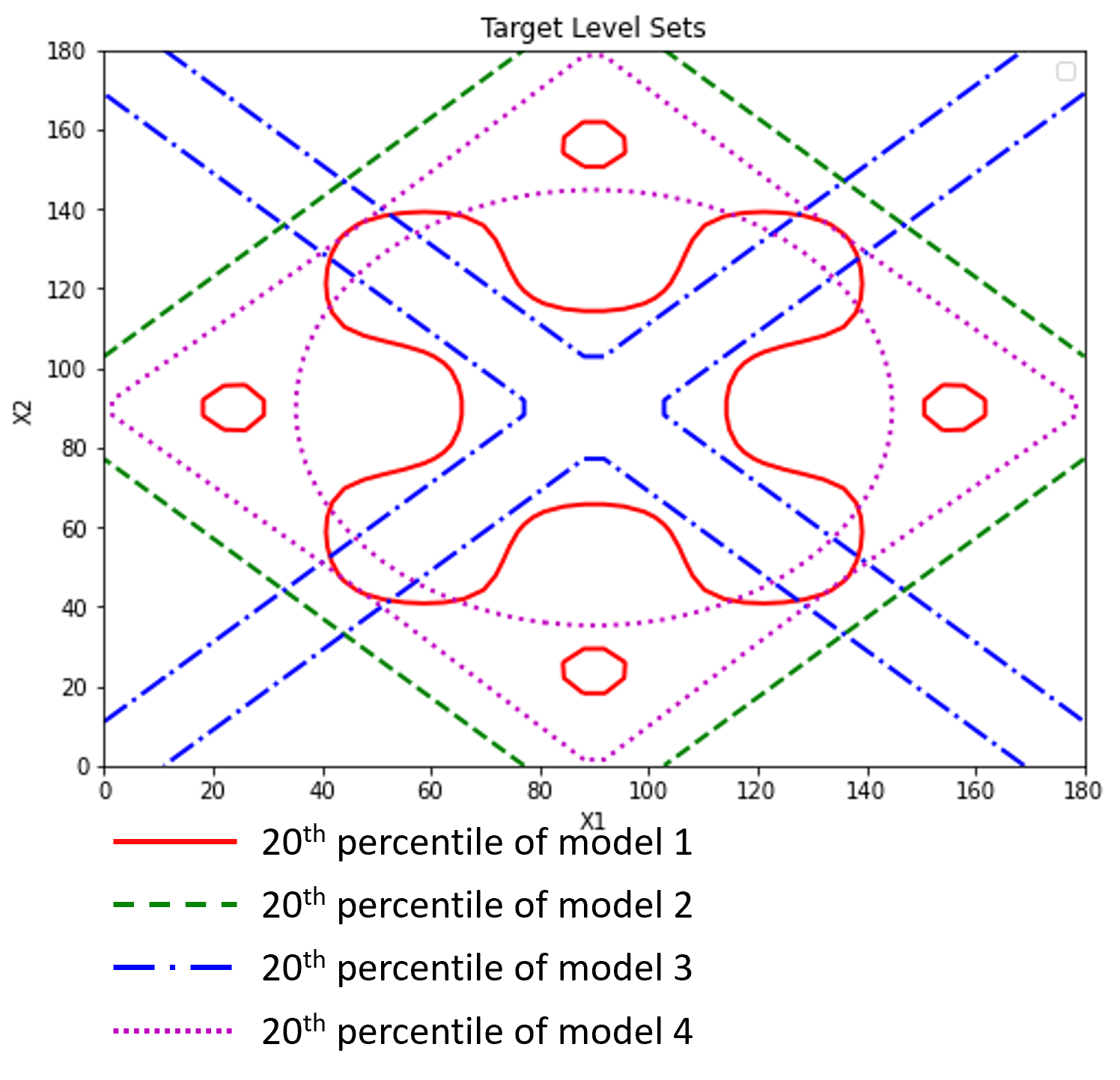}
    \caption{Target regions of each model in Example~2. }
    \label{fig:levelsets2}
\end{subfigure}
\hfill
\begin{subfigure}{.47\textwidth}
    \centering
    \includegraphics[width=0.99\linewidth]{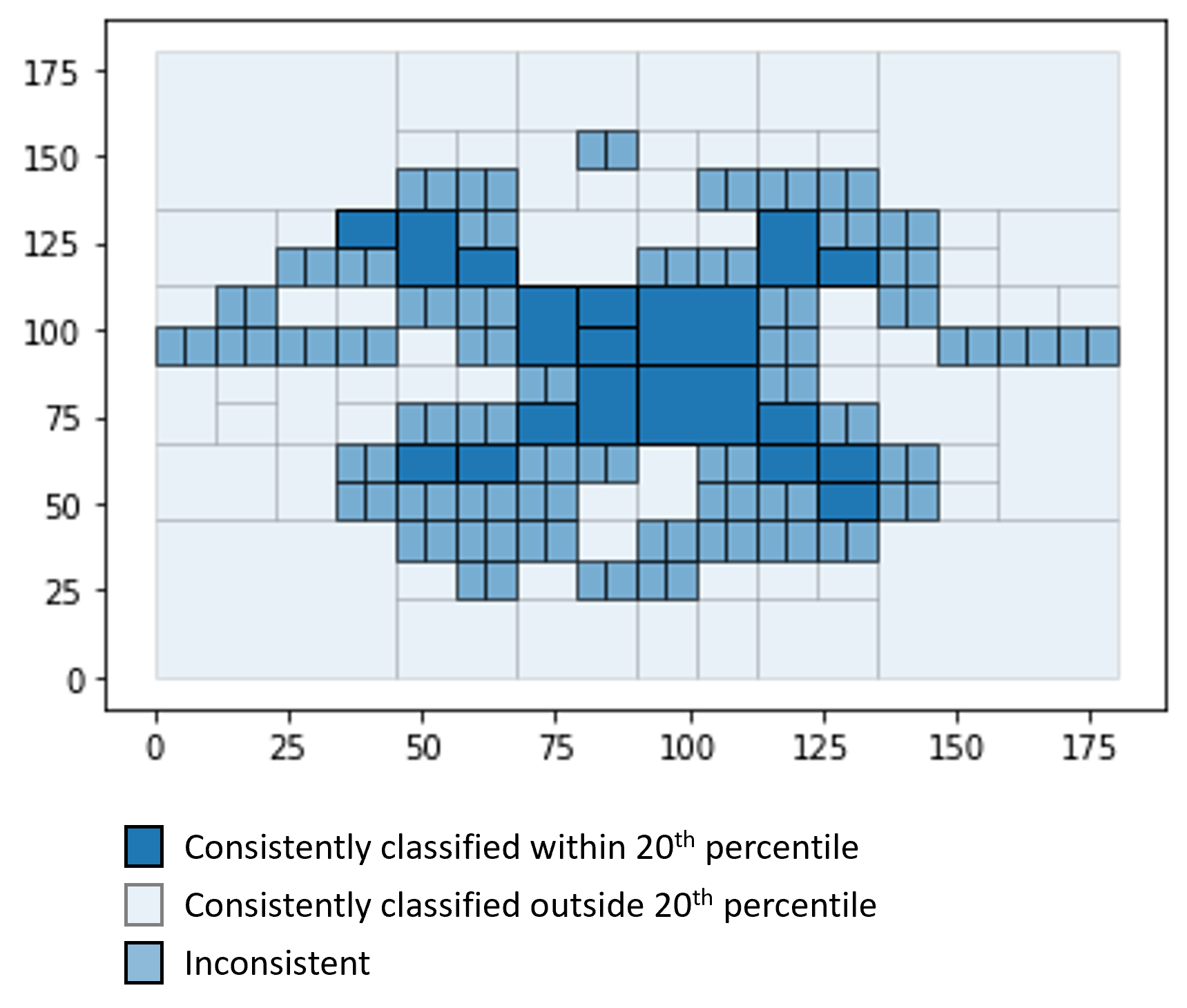}
    \caption{Example 2 results after termination at 10 iterations. }
    \label{fig:empiricalresults2}
\end{subfigure}
\caption{Example 2 target regions (a) and solution set (b).}
\end{figure}

After 10 iterations, S-BOMM consistently classified approximately $11\%$ of the region as within the 20th percentile target region. These results are shown in Figure~\ref{fig:empiricalresults2}. 
A large portion of the regions that are consistently classified within the 20th percentile reside around the center of the feasible region. This is because most of the models  agree that this region is within the 20th percentile. There are large subregions that are consistently classified as outside of the 20th percentile where all or most models agree. S-BOMM successfully located a set of consistently ``good" subregions, even with non-convex target regions.

\subsection{\emph{Discussion}}\label{s:discussion}
The empirical results demonstrate that S-BOMM is capable of identifying consistently desirable regions even when the target regions according to different models are non-convex. 
Because the method relies on agreement across models, it remains applicable in settings where other optimization approaches with multiple models may be difficult to apply directly.

It should be noted that a limitation of the methodology is its reliance on sufficient overlap among the target regions of the models in order for consistency to be established. The target regions of the three models in Example 1 (Section~\ref{s:example1}) exhibit large amounts of overlap, leading to sufficient consistent classification after 10 iterations with $v=1$ and $r=1$. The target regions of the four models in Example 2 (Section~\ref{s:example2}) exhibit much less overlap. In order to produce sufficient consistent classification after 10 iterations, the parameter $r$ had to be reduced to $r=0.75$. This behavior illustrates that the degree of agreement among models directly influences the parameter choices and resulting behavior of S-BOMM, reinforcing that the method is most effective when the models share at least moderate consistency in their characterization of desirable regions.

Finally, the empirical studies align with known properties of the models. For example, Model 3 of Example 1, as given in (\ref{eq:mod3}), is a Taylor series approximation centered at the origin and is therefore expected to be more reliable near that point and less reliable farther away. The consistently classified regions occur in areas where this approximation is most credible relative to the other models, providing an interpretable explanation for the observed results without requiring any model to be treated as authoritative.

\section{Conclusion}\label{s:conclusion}
This paper presents S-BOMM, a methodology designed for set-based optimization of complex systems  in which multiple models provide differing perspectives on a complex system, but a single, most accurate, high-fidelity model is unavailable or impractical to determine. 
We introduce a consistency score and the concept of consistent classification among models that are embedded in S-BOMM.
We demonstrate the capabilities of S-BOMM through an analysis of the probability of correct and incorrect consistent classification. Additionally, we demonstrate S-BOMM through two empirical examples and discuss practical recommendations for implementation.
In doing so, we have shown that S-BOMM provides a framework for optimization in settings where no reliable hierarchy of model accuracy exists.

The S-BOMM methodology is most effective when at least three models are available.
In such cases, S-BOMM leverages regions of consensus to guide decision making without requiring a single high-fidelity model.
However, if too many models provide substantially different characterizations of the system, achieving meaningful consistency may become difficult. The similarity of models should also be considered. For example, if there are several very similar models (e.g., structurally similar simulations with different input conditions) and only one or a few models of a different perspective (e.g., a queueing model), agreement among the similar models may disproportionately influence consistency. In this way, careful consideration should be given to ensure a balance of different model perspectives.

The methodology is also well suited to applications where expert judgment plays an essential role. 
Because S-BOMM yields a set of ``good" solutions rather than a single prescribed optimum, it aligns well with decision environments where flexibility, interpretability, and adaptability are valued over precision of a single solution.

Parameters $v$ and $r$ in S-BOMM, which govern the thresholds for consistent classification, depend not only on the decision maker’s tolerance for risk, but also on the confidence in the underlying model classifications.
By definition, the values for both $v$ and $r$ are bounded above by $N$, the number of models. However, both parameters should remain strictly less than $N$, since setting either equal to $N$ would require unanimous agreement with perfect classification confidence, which is not practical in real-world settings.
Smaller values for $v$ and $r$ will more readily result in consistent classification, but may yield higher error, whereas larger $v$ or $r$ may take significant computational expense to yield consistent classification.  In practice, determining appropriate values for $v$ and $r$ is highly application-specific and may require some degree of trial and error.
If few regions are consistently classified after many iterations, decreasing either $v$ or $r$ can promote greater flexibility, resulting in consistent classification. Conversely, if very large regions are classified consistently after only a few iterations, larger values of $v$ and/or $r$ may be warranted to avoid premature termination. Situations in which models capture similar features or measure related aspects of the system may also justify larger values of $v$ and $r$, since greater agreement is expected among the models.

Several avenues for future research remain to further strengthen and generalize the {S-BOMM} framework. Potential directions include examining how alternative sampling and classification schemes influence performance. Additionally, the current analysis assumes independence among models, an assumption that may not hold in practice when models share data sources or when one model is constructed as a surrogate of another. Investigating the impact of correlated models and identifying strategies to mitigate any resulting performance degradation is another area of future study. Further work could also explore how characteristics of the models themselves affect S-BOMM outcomes, such as the effects of systematic translations, noise, or other structural variations, to better understand the method’s sensitivity and robustness across diverse multiple model settings.

\if0\blind{
\section*{Acknowledgements}
This research has been funded in part by the Office of Naval Research and the National Science Foundation. Generative AI (GPT-5.2) was used for the purposes of language improvement and coding assistance.} \fi


\bibliographystyle{chicago}
\spacingset{1}
\bibliography{IISE-Trans}

\end{document}